\documentclass[12pt]{article}
\usepackage{graphicx}
\usepackage{natbib} 
\usepackage{url} 

\usepackage{amssymb}
\usepackage{bm}
\usepackage{amsmath}
\usepackage{algorithm}
\usepackage{algorithmic}
\usepackage{multirow}
\usepackage{color}
\usepackage{enumerate}
\usepackage{rotating}

\newcommand{\blind}{0}

\begin{document}

\def\spacingset#1{\renewcommand{\baselinestretch}%
{#1}\small\normalsize} \spacingset{1}


\if0\blind
{
  \title{\bf Online Structural Change-point Detection of High-dimensional Streaming Data via Dynamic Sparse Subspace Learning}
  \author{Ruiyu Xu \\
    Department of Industrial Engineering and Management\\ Peking University\\
    and \\
    Jianguo Wu \\
    Department of Industrial Engineering and Management\\ Peking University\\
    and \\
    Xiaowei Yue \\
    Department of Industrial \& Systems Engineering\\ Virginia Tech\\
    and \\
    Yongxiang Li \\
    Department of Industrial Engineering and Management\\ Shanghai Jiao Tong University}
 \date{}
  \maketitle
} \fi

\if1\blind
{
  \bigskip
  \bigskip
  \bigskip
  \begin{center}
    {\LARGE\bf Online Structural Change-point Detection of High-dimensional Streaming Data via Dynamic Sparse Subspace Learning}
\end{center}
  \medskip
} \fi

\spacingset{1.5}

\bigskip
\bigskip\bigskip
\begin{abstract}
High-dimensional streaming data are becoming increasingly ubiquitous in many fields. They often lie in multiple low-dimensional subspaces, and the manifold structures may change abruptly on the time scale due to pattern shift or occurrence of anomalies. However, the problem of detecting the structural changes in a real-time manner has not been well studied. To fill this gap, we propose a dynamic sparse subspace learning approach for online structural change-point detection of high-dimensional streaming data. A novel multiple structural change-point model is proposed  and the asymptotic properties of the estimators are investigated. A tuning method based on Bayesian information criterion and change-point detection accuracy is proposed for penalty coefficients selection. An efficient Pruned Exact Linear Time based algorithm is proposed for online optimization and change-point detection. The effectiveness of the proposed method is demonstrated through  several simulation studies and a real case study on gesture data for motion tracking. 
\end{abstract}

\noindent%
\vfill

\newpage
\spacingset{1.5} 
\section{Introduction}
\label{sec1}
High-dimensional streaming data are ubiquitous in many fields such as bioinformatics, engineering, finance and social sciences. For example, in biological studies, neurons being monitored could generate hundreds or thousands of time series signals \citep{qiu2016joint}. In image signal processing, each dynamic image in a high resolution video could consist of more than one hundred thousand pixels. In semiconductor manufacturing, hundreds of sensors are installed in the production system for real-time monitoring of the manufacturing condition \citep{zhang2020dynamic}. In gesture tracking, tens of sensors are mounted to dynamically capture the positions of body joints \citep{jiao2018subspace}. The relationship or correlation among these dimensions is of great value for research, as it provides insights into regularities and inter-dependencies between observed variables \citep{kolar2012estimating}. Usually, the correlation or dependence structure is sparse, i.e., a variable is only correlated with a small proportion of other variables. Besides, the dependence structure may change over time and the change-points often imply events or anomalies occurring at that moment. For instance, changes in the correlation between brain nerves may represent shifts in thinking patterns \citep{haslbeck2015mgm} or the onset of seizure \citep{dhulekar2015seizure}. Changes in the correlation between image pixels may indicate transitions in the subject of the video \citep{tierney2014subspace}.  In tunnel excavation process, the torque of the cutterhead of the Tunnel Boring Machine (TBM) has a linear relationship with the penetration rate, and the change in the regression coefficient may result from the change of geological condition \citep{shi2019fuzzy}. Therefore, online detection of change in correlation or inter-dependence is of great importance to determine whether an event or anomaly has recently occurred in the system.

In this paper, we refer to the change in  linear relationship as a structural change,  i.e., change of the manifold structure capturing the linear relationship among the high-dimensional streaming data. The structural change-points separate the multivariate streaming data into multiple segments, where each segment has a unique structure modeling their relationship. Multiple change-point problems have been actively studied in many fields, e.g., economics, climatic time series \citep{aminikhanghahi2017survey, wu2016online, wu2019sequential}. However, these problems often refer to detection of breaks in trend or distributional parameters, e.g., a shift in mean or variance, while structural change detection focuses on detecting the changes of the underlying relationships among different dimensions. The structural change-point detection problem, especially the online one, has not been well explored compared with the traditional multiple change-point detection problems. Due to the “curse of dimensionality” \citep{bellman1966dynamic}, it is often challenging to detect these change-points accurately and timely  for high-dimensional streaming data. Too many variables constitute an extremely complex correlation structure that is hard to estimate. Noise contamination and insufficient sample size further increase the difficulty of estimation.

The Gaussian graphic model (GGM, \citealp{dempster1972covariance}) is a widely used method and continues to attract much attention to study the inter-dependence structure of multiple variables. A common assumption is that the sample $X \sim N_{ p}(0, \Sigma)$ is a $ p$-dimension Gaussian vector. Let $\Omega:=\Sigma^{-1}$ denote the precision matrix, with entries $\left(\omega_{i j}\right), 1 \leq i, j \leq  p$. It can be easily shown that the precision matrix $\Omega$ encodes the conditional independence structure among the variables. Variable $i$ and $j$ are conditionally independent given all other coordinates of $X$ if and only if the entry $\omega_{i j}$ of the precision matrix is zero. \citet{meinshausen2006high} were the first to combine GGM with LASSO to get a sparse precision matrix $\Omega$, and later a more systematic approach named Graphic LASSO was proposed by \citet{tibshirani2005sparsity}. Since then, there has been much similar work on estimating a single precision matrix $\Omega$ based on $n$ independent samples \citep{drton2008sinful, foygel2010extended, rothman2008sparse, yuan2006model, ren2015asymptotic}.  In addition, some Gaussian graphical models for functional data are also developed to model the correlations of the functions \citep{qiao2019functional, qiao2020doubly, gomez2020functional,zhu2016bayesian,li2018nonparametric}. However, these methods cannot dynamically track the change of the relationships or dependency graphs over time. To this end, several research groups \citep{kolar2011time, haslbeck2015mgm, qiu2016joint, zhou2010time} assumed that the dependency graph evolves continuously over time and proposed kernel smoothing methods for estimating time-varying graphical models. \citet{kolar2012estimating} assumed that the graph changes abruptly at some time instants and proposed a penalized neighborhood selection method with a fused-type penalty for estimating a piece-wise constant graphical model. Considering that there may be prior knowledge of potential groups, \citet{gibberd2017regularized} proposed a group-fused graphical lasso estimator for grouped estimation of change-points.  However, all the methods above model the data with multivariate Gaussian distributions with constant means, which may not be applicable to streaming data with means continuously changing in a manifold structure.

Sparse subspace clustering (SSC) is another type of methods that can be used to capture the sparse dependencies or correlations across different variables \citep{elhamifar2013sparse}. Subspace clustering is an extension of traditional clustering that seeks to find clusters in different subspaces. It is based on the fact that high-dimensional data often lie in multiple subspaces of significantly lower dimension instead of being uniformly distributed across the full space \citep{parsons2004subspace}. The key idea of SSC is the self-expressive property with sparse representation, i.e., each data point in a union of subspaces can be sparsely represented as a linear or affine combination of other points from its own subspace. SSC method builds a similarity graph by these sparse coefficients, and obtains data segmentation using spectral clustering. Later several structured SSC were developed by integrating the two separate stages of computing a sparse representation matrix and applying spectral clustering into a unified optimization framework \citep{li2015structured, li2017structured, zhang2016low}. \citet{tierney2014subspace} proposed an ordered subspace clustering method by including a new penalty term to handle data from a sequentially ordered union of subspaces. \citet{guo2013spatial} proposed a spatial subspace clustering (SpatSC) by combining subspace learning with the fused lasso for 1D hyperspectral data segmentation. However, all of these methods focus on static data of fixed length, and thus are not applicable to dynamic streaming data with increasing length. Besides, they are not able to detect the dynamic change of cross-correlation structure among variables. Recently, \citet{zhang2020dynamic} proposed a dynamic multivariate functional data modeling approach to capture the change of cross-correlation structure over time. By formulating the problem as a sparse regression with fused LASSO penalty, the correlation structure among different variables as well as the change-points can be efficiently estimated using the Fast Iterative Shrinkage-thresholding Algorithm (FISTA). Nevertheless, this method is offline and cannot sequentially estimate the cross-correlation structure and detect the change-points. \citet{jiao2018subspace} proposed an online cumulative sum (CUSUM)-based control chart for subspace change-point detection. This method first learns the pre-change subspace from historical data, and then conducts online detection via a CUSUM statistic.  However, this method is limited to one change-point scenarios.  Besides, it requires sufficient historical pre-change data to get the basis of the subspace.

To fill the research gap, we propose a novel dynamic sparse subspace learning  (DSSL) approach for online detecting the change of sparse correlation structure of high-dimensional streaming data. Specifically, we follow the self-expressive assumption in \citep{elhamifar2013sparse} and formulate a multiple structural change-point model with two penalty terms in the loss function for encouraging sparse representation and avoiding excessive change-points respectively.  We further derive several asymptotic properties for the model estimators, showing that the positions of the change-points converge to the true values as the length of segments increases, and the estimated coefficients satisfy a LASSO subspace detection property. A PELT \citep{killick2012optimal} based dynamic programing algorithm is developed for online model optimization and change-point detection.

The rest of this paper is organized as follows.  In Section \ref{sec2}, two basic assumptions for sparse subspace learning are given, and a multiple structural change-point model is formulated. The asymptotic properties of the proposed model are investigated in Section \ref{sec3}. In Section \ref{sec4}, we show how to solve the optimization problem sequentially via  a customized PELT algorithm, and propose some strategies to determine the penalty coefficients and proper hyperparameters to improve the computational efficiency. In Section \ref{sec5}, numerical experiments with synthetic and real data are conducted to demonstrate and validate the effectiveness of the proposed algorithm. Section \ref{sec6} presents the discussions and conclusions.

\section{Multiple Structural Change-point Modeling via Dynamic Sparse Subspace Learning}
\label{sec2}
In this section, we first  illustrate the structural change and the motivation of using dynamic sparse subspace learning through a simple example, and then introduce the subspace assumption and the self-expressive assumption  (property), which lay the foundation for SSC.  After that, a multiple structural change-point modeling approach is formulated for change detection. 
 
\subsection{Illustration of the Structural Change}
\begin{figure}[htb]
  \centering
  \includegraphics[width=\textwidth]{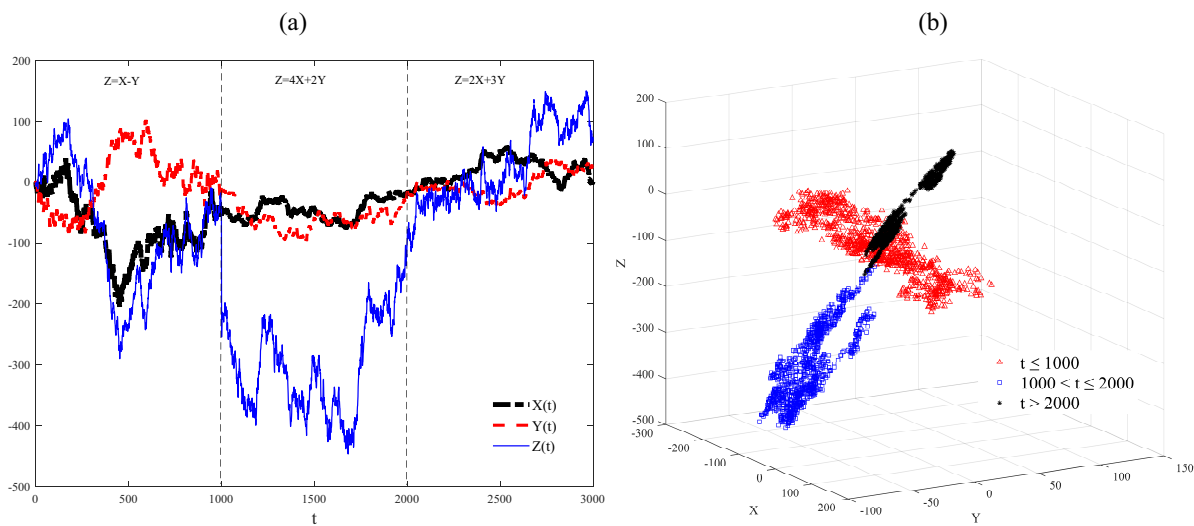} 
  \caption{ Illustration of the structural change.}
  \label{fig0} 
\end{figure}

As mentioned earlier, the structural change here is defined as the change of linear manifold capturing the relationship among the times series of the streaming data. Figure \ref{fig0} is a simple example with three time series illustrating the concept of structural change. In Figure 1(a), there are two change-points, i.e., $t=1000$ and $2000$. Once a change-point occurs, the linear relationship among these variables changes, e.g., from $Z=X-Y$ to $Z=4X+2Y$ at $t=1000$. Figure 1(b) shows the time series in a three-dimensional space. Clearly, each segment in Figure 1(a) corresponds to a plane or a linear manifold in Figure 1(b), and we can easily see the transition from one to another.  

In practical applications with high-dimensional streaming data, however, a variable may be related with only a small number of other variables, yielding multiple clusters of much lower dimensions, which makes the subspace learning very challenging. To overcome this issue, we borrow the idea of SSC and develop a dynamic subspace learning approach. The technical details are given in the following subsections.

\subsection{Notations and Basic Assumptions}
\label{sec21}
Consider a $p$-dimensional (e.g., $p$-channels) streaming data $\left(\boldsymbol{Y}_{1}, \ldots, \boldsymbol{Y}_{p}\right)'$, where each dimension is of length $N$ on the time scale, e.g., $\boldsymbol{Y}_{i}=\left(Y_{i 1}, Y_{i 2}, \ldots, Y_{i N}\right)'$. We assume 
\begin{equation}Y_{i j}=X_{i j}+\epsilon_{i j},\ i=1, \ldots, p,\ j=1, \ldots, N,\end{equation}
where $X_{i j}$ is the true value and $\epsilon_{i j}$ is the independent noise with mean $E\left[\epsilon_{i j}\right]=0$ and variance $\operatorname{Var}\left[\epsilon_{i j}\right]=\sigma_{0}^{2}$. We assume there is no autocorrelation in the noise. To facilitate understanding, we could treat a discrete time series $\boldsymbol{X}_{i}$ as a functional sample of ${X}_{i}(t)$, and assume that these functions can be partitioned into $L$ different subspaces $S_{l}, l=1, \ldots, L$. Functions in the same subspace have strong cross-correlations, while functions in different subspaces have no cross-correlations. To facilitate subspace learning, we present two required assumptions, which are similar to Zhang's work \citep{zhang2020dynamic}.
\\

\noindent
{\bf Assumption A1.} (Subspace Assumption)
{\it It is assumed that the multivariate streaming data can be partitioned into different subspaces. Each subspace or translated subspace $S_{l}$ is defined as the set of all functions formed by linearly combining the $d_{l}$ basis functions $\boldsymbol{\Phi}_{l}=\left(\phi_{l 1}(t), \ldots, \phi_{l d_{l}}(t)\right)'$ with a translation or shift function $\phi_{l 0}(t)$
\begin{equation}\boldsymbol{S}_{l} \triangleq\left\{X(t) \mid X(t)=\sum_{q=1}^{d_{l}} a_{q} \phi_{l q}(t)+\phi_{l 0}(t), a_{q} \in \mathcal{R}\right\}.\end{equation}
If orthogonal basis functions are considered, then,
$$\int \phi_{l i}(t) \phi_{l j}(t) d t=0, \text{ for } i, j=1, \ldots, d_{l}, i \neq j.$$
} 
Note that if $\phi_{l 0}(t) \neq 0$, it is a translated subspace, as it does not contain the origin \citep{nowinski1981applications}.  
This subspace assumption tells us that the time series within the same subspace share common basis functions, thus are expected to have strong correlations or linear relationship. This assumption is reasonable in many cases. For example, in EEG signals \citep{thatcher2005eeg}, the collected signals in a certain functional zone of the brain may be highly correlated, as they are the superposition of electrical signals from the same set of neurons. The activity of these neurons form the shared basis functions. As there are many functional zones in the brain, the EEG signals can be partitioned into multiple subspaces. Similar phenomenon can also be found in seismic signals caused by multiple earthquake events \citep{kitov2019use}, and river water flows within the same region and across multiple regions.

In practical applications, the collected streaming data are discrete. In such a case, the basis vectors, e.g., $\boldsymbol{\phi}_{l i}=\left(\phi_{l i}\left(t_{1}\right), \ldots, \phi_{l i}\left(t_{N}\right)\right)$ instead of the basis functions can be used to represent the subspace assumption. The formed vector subspace is actually an affine space.  
With this assumption, we have $\boldsymbol{X}=\boldsymbol{A} \boldsymbol{\Phi}$ where $\boldsymbol{X}=\left(\boldsymbol{X}_{1}, \ldots, \boldsymbol{X}_{p}\right)'$ is the true-value time series data in $\mathcal{R}^{p \times N}$, $\boldsymbol{\Phi}=\left(\boldsymbol{\phi}_{11}, \ldots, \boldsymbol{\phi}_{1,d_1},\ldots,\boldsymbol{\phi}_{L,d_L}\right)'$ is the discrete basis matrix in $\mathcal{R}^{d \times N},\ d=\sum_{l=1}^{L}{d_l}$ and $\boldsymbol{A}$ is a sparse coefficient matrix in $\mathcal{R}^{p \times d}$ with entry $\alpha_{i j}\neq0$ if and only if $\boldsymbol{X}_{i}$ belong to the subspace $\boldsymbol{S}_{l}$ which contains the basis function $\boldsymbol{\Phi}_{j}$.

\begin{figure}[htb]
  \centering
  \includegraphics[width=.6\textwidth]{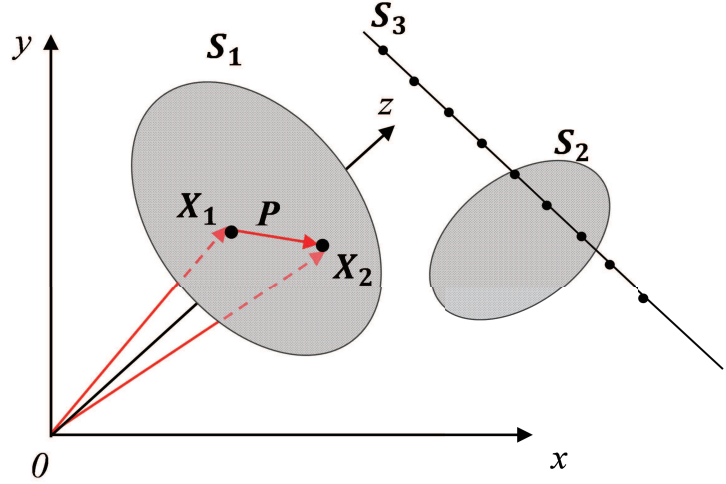} 
  \caption{Illustration of one-dimensional ($S_{3}$) and two-dimensional ($S_{1}$ and $S_{2}$) subspaces in ${\mathcal{R}}^{3}$.} 
  \label{fig1} 
\end{figure}

  Note that here the dimensionality of the space is the length of the time series $N$ (row space of $\boldsymbol{X}$), not the number of time series $p$ (column space). Each time series here represents a point, with $p$ points in total. This representation is different from Figure 1(b) where the column space is used. In column space, there is only one subspace, while in row space, there may be multiple such spaces, with each one corresponding to a cluster of the time series. Figure \ref{fig1} is an illustrative example showing three subspaces in ${\mathcal{R}}^{3}$. Intuitively, given sufficient time series, each time series can be efficiently represented as a linear combination of other time series in the same subspace. For example, in subspace $S_{1}$, point $\boldsymbol{X}_{2}=\boldsymbol{X}_{1}+\boldsymbol{P}$ where $\boldsymbol{P}=\alpha_{1}\left(\boldsymbol{X}_{4}-\boldsymbol{X}_{3}\right)+\alpha_{2}\left(\boldsymbol{X}_{6}-\boldsymbol{X}_{5}\right)$ for certain points $\boldsymbol{X}_{3},\boldsymbol{X}_{4},\boldsymbol{X}_{5}$ and $\boldsymbol{X}_{6}$ in $S_{1}$. This self-expressive property is summarized in Assumption A2 as follows.

\noindent
{\bf Assumption A2.} (Self-Expressive Assumption)
{\it If there are sufficient  time series} from each subspace, e.g., $p_{l}>d_{l}$ for $l=1, \ldots L$, where $p_{l}$ is the number of time series in subspace $S_{l}$, we have and $d_{l}$ is the subspace dimension, then  $\boldsymbol{X}_{i}=\left(\boldsymbol{X}_{i1},\ldots,\boldsymbol{X}_{iN}\right)'$ is self-expressive, i.e., for all $ i \in \boldsymbol{P}_{l}$ where $\boldsymbol{P}_{l}$ is the set of time series indices of subspace $S_{l}$, we have
\begin{equation}\boldsymbol{X}_{i}=\sum_{j \in \boldsymbol{P}_{l}, j \neq i} \beta_{i j} \boldsymbol{X}_{j}, \text{ for } i=1, \ldots, p.\end{equation}

With this assumption, we have $\boldsymbol{X}=\boldsymbol{B} \boldsymbol{X}$ and $\boldsymbol{A}=\boldsymbol{B} \boldsymbol{A}$, where $\boldsymbol{X}=\left(\boldsymbol{X}_{1}, \ldots, \boldsymbol{X}_{p}\right)'$ is the noise-free time series data in $\mathcal{R}^{p \times N}$, $\boldsymbol{A}$ is the sparse coefficient matrix of basis functions in $\mathcal{R}^{p \times d}$, $\boldsymbol{B}$ is a sparse coefficient matrix in $\mathcal{R}^{p \times p}$ with entry $\beta_{i i}=0$ and $\beta_{i j}=0$ if $\boldsymbol{X}_{i}$ and $\boldsymbol{X}_{j}$ belong to different subspaces. This self-expressive property can be easily derived based on Assumption A1. A detailed discussion is provided in the Supplementary Materials.

Note that the coefficient matrix $\boldsymbol{B}$ is not unique and could have infinite solutions. Naturally, an optimal $\boldsymbol{B}$ can be obtained by minimizing the objective function with the $l_q$-norm of the solution, that is,
\begin{equation}argmin_{\boldsymbol{B}} \ ||\boldsymbol{B}||_{q} \  \text{s.t.} \ \boldsymbol{X}=\boldsymbol{B} \boldsymbol{X} ,\ \beta_{i i}=0, \text{ for } i=1,\dots,p.\nonumber\end{equation}
As $q$ decreases from 1 to 0, the sparsity of $\boldsymbol{B}$ increases. Although a sparsest $\boldsymbol{B}$ could be obtained with $l_0$-norm, the problem turns to a NP-hard problem, which is difficult to solve. As an alternative, $l_1$-norm is widely used with good sparsity property and computational efficiency. It is obvious that this solution has sparsity between subspaces, that is, $\beta_{i j}=0$ if $\boldsymbol{X}_{i}$ and $\boldsymbol{X}_{j}$ belong to different subspaces.

\subsection{Model Formulation}
\label{sec23}
The self-expressive property states that each time series can be efficiently reconstructed by a linear combination of other time series in the same subspace. Considering the existence of measurement noise, the sparse representation $\boldsymbol{B}$ can be obtained by minimizing the following objective function with an $l_{1}$ penalty
 
\begin{equation}
\label{equ5}
\min _{\beta_{i j}, j \neq i} \sum_{i=1}^{p}\left\{\frac{1}{2}\sum_{t=1}^{N}\left(Y_{i t}-\sum_{j \neq i} \beta_{i j} Y_{j t}\right)^{2}+\lambda_{1} \sum_{j \neq i}\left|\beta_{i j}\right|\right\},\ \text {s.t.}\ \beta_{i i}=0,\  \text {for}\  i=1, \ldots, p,\end{equation}
where $\lambda_{1}$ is the penalty weight to control the sparsity in the representation. As the representation of each time series is independent of those of other time series, the above optimization problem can be efficiently solved efficiently using the LASSO algorithm.

In the traditional sparse subspace clustering, the dimension of the data points and the correlation structure are fixed. However, in our case, the length of the streaming data dynamically increases and there may be abrupt structural changes, i.e., changes in $\beta_{i j}, j \neq i$ at some unknown change-points. For a $p$-channel streaming data of length $N$, suppose there are in total $C$ change-points $\left\{\tau_{1}, \ldots, \tau_{C}\right\}$ with $0<\tau_{1}<\cdots<\tau_{C}<N$, which partitions the streaming data into $C+1$ segments. For notational convenience, we define $\tau_{0}=0$ and $\tau_{C+1}=N$. In the estimation of change-point models, the fused lasso is one of the most popular techniques, which penalizes the $l_{1}$-norm of both the coefficients and their successive differences \citep{tibshirani2005sparsity, tierney2014subspace, zhang2020dynamic}. Using the fused lasso, the problem can be formulated as
\begin{equation}
\label{equ6}
\min _{\beta_{i j t}, j \neq i, \atop t=1, \ldots, N} \sum_{t=1}^{N}\left[\frac{1}{2}\left(Y_{i t}-\sum_{j \neq i} \beta_{i j t} Y_{j t}\right)^{2}+\lambda_{1} \sum_{j \neq i}\left|\beta_{i j t}\right|\right]+\lambda_{2} \sum_{t=2}^{N} \sum_{j \neq i}\left|\beta_{i j t}-\beta_{i j t-1}\right|, i=1, \ldots, p.
\end{equation}

The above optimization can be achieved by the Fast Iterative Shrinkage-thresholding Algorithm (FISTA, \citealp{beck2009fast}). However, this formulation is inherently offline and cannot be efficiently solved in a real-time or sequential manner for online applications. To overcome this problem, we propose a new model formulation that can be sequentially optimized

\begin{gather}
\label{equ7}
\min _{C, \tau_{1}, \ldots, \tau_{C} \atop \boldsymbol{B}^{(c)}, c=1, \ldots, C+1} \sum_{c=1}^{C+1}\left\{\sum_{i=1}^{p}\left[\sum_{t=\tau_{c-1}+1}^{\tau_{c}} \frac{1}{2}\left(Y_{i t}-\sum_{j \neq i} \beta_{i j}^{(c)} Y_{j t}\right)^{2}+ {\lambda_1(\delta^{(c)})} \sum_{j \neq i}\left|\beta_{i j}^{(c)}\right|\right]+\lambda_{2}\right\},\\ \text{s.t. } \beta_{ii}^{(c)}=0, \nonumber
\end{gather}

or
\begin{gather}
\min _{C, \tau_{1}, \ldots, \tau_{C} \atop \boldsymbol{B}^{(c)}, c=1, \ldots, C+1} \sum_{c=1}^{C+1} \sum_{i=1}^{p}\left(\frac{1}{2}\left\|\boldsymbol{Y}_{i}^{(c)}-\boldsymbol{\beta}_{i}^{(c)}\boldsymbol{Y}^{(c)} \right\|_{2}^{2}+ {\lambda_1(\delta^{(c)})} \left\|\boldsymbol{\beta}_{i}^{(c)}\right\|_{1}\right)+\lambda_{2}(C+1),\\
\text{s.t. } \beta_{ii}^{(c)}=0, \nonumber
\end{gather} 
where $\boldsymbol{Y}^{(c)}=\left(\boldsymbol{Y}_{1}^{(c)}, \ldots, \boldsymbol{Y}_{p}^{(c)}\right)'$ is the streaming data in the $c$th segment, $\boldsymbol{Y}_{i}^{(c)}=\\ \left(Y_{i \tau_{c-1}+1}, \ldots, Y_{i \tau_{c}}\right)^{\prime}$, $ \boldsymbol{\beta}_{i}^{(c)}=\left(\beta_{i 1}^{(c)}, \beta_{i 2}^{(c)}, \ldots, \beta_{i p}^{(c)}\right)$, $ \beta_{i i}^{(c)}=0$ are the representation coefficients for channel $i$ in the $c$th segments, $\boldsymbol{B}^{(c)}=\left(\boldsymbol{\beta}_{1}^{(c)}, \ldots, \boldsymbol{\beta}_{p}^{(c)}\right)$, $\lambda_1(\delta^{(c)})$ is a sparsity penalty which may dynamically change with the length of the current segment $\delta^{(c)}$, and $\lambda_{2}$ is a penalty weight penalizing the number of segments to avoid overfitting. 

The formulation of Equation (\ref{equ7}) which explicitly incorporate the change-points in the model has several advantages compared with Equation (\ref{equ6}). First of all, the number of parameters significantly decreases by directly setting constant representation coefficients in each segment, which greatly reduces the problem complexity. Secondly, the smoothness penalty term $\lambda_{2} \sum_{t=2}^{N} \sum_{j \neq i}\left|\beta_{i j t}-\beta_{i j t-1}\right|$ in Equation (\ref{equ6}) tends to reduce the differences between two successive segments, while Equation (\ref{equ7}) does not have such an issue. 
Thirdly and most importantly, Equation (\ref{equ7}) can be sequentially solved  via an efficient dynamic programming approach with controlled computational cost and low memory requirement, which will be shown in detail in Section 4. However, for the Equation (\ref{equ6}), we need to restart the whole optimization process from the very beginning once a new observation arrives. The rapid growth of the computational cost as well as the memory requirement makes it very prohibitive in online applications.  We name our formulation along with the sequential optimization algorithm as dynamic sparse subspace learning (DSSL).

\section{Asymptotic Properties}
\label{sec3}
In this section, the asymptotic properties of the estimators of DSSL are established.  We first discuss the asymptotic properties of the optimal solution $\boldsymbol{B}$ of equation (\ref{equ5}) with no change-points. The convergence property and sparsity property are proved as the sample size $N_0$ goes to infinity. Then, we prove that, as the sample size $N$ of each segment goes to infinity, the locations of the change-points could be precisely detected and the estimated coefficient matrices $\hat{\boldsymbol{B}}^{(c)}$ of each segment satisfy the LASSO Subspace Detection Property introduced by \citep{wang2013noisy}, a very desirable property for subspace learning. For convenience, we provide the definition of the LASSO Subspace Detection Property as follows.\\

\noindent
{\bf Definition 1 (LASSO Subspace Detection Property)}\\
The estimated coefficient matrices $\hat{\boldsymbol{B}}$ is called satisfying the LASSO Subspace Detection Property if for all $i=1,\ldots,p,$\\
(1). $\hat{\boldsymbol{\beta}}_{i}$ is not a zero vector, i.e., the solution is non-trivial, \\
(2). Nonzero entries of $\hat{\boldsymbol{\beta}}_{i}$ correspond to only the time series sampled from the same subspace as $\boldsymbol{Y}_{i}$.\\

For simplicity yet without loss of generality, we only consider one sparse regression in the following analysis, e.g., $\boldsymbol{Y}_{i}$ as the response variable. The problem of including all sparse regressions can follow the same approach.
 
Denote the $\boldsymbol{A}_{i} \in \mathcal{R}^{1 \times d}$ as the $i$-th row of $\boldsymbol{A}$ and the $\boldsymbol{A}_{-i} \in \mathcal{R}^{(p-1) \times d}$ as the rest rows.\\

%

%

\noindent
{\bf Theorem 1.} \\
{\it Under (A1-A2), for the optimization problem {\textbf{without}} change-points
\begin{equation}
\label{equ8}
\min _{\beta_{i j}, j \neq i} \frac{1}{2}\sum_{t=1}^{N_0}\left(Y_{i t}-\sum_{j \neq i} \beta_{i j} Y_{j t}\right)^{2}+\lambda_1(N_0) \sum_{j \neq i}\left|\beta_{i j}\right|,\ \text {s.t.}\ \beta_{i i}=0,\end{equation}
where $\lambda_1(N_0)$ is a function of sample size $N_0$,\\
(1). When $\lambda_1(N_0) /N_0 \rightarrow 0$, the estimated $\hat{\boldsymbol{\beta}}_{i}$ of (\ref{equ8}) converges to a sparse vector\\ $\boldsymbol{A}_{i}\boldsymbol{A}'_{-i} \left(\boldsymbol{A}_{-i}\boldsymbol{A}'_{-i}+\sigma^2_0\boldsymbol{I_{p-1}}\right)^{-1}$ as $N_0$ goes to infinity.\\
(2). If $\lambda_1(N_0) /N_0 \rightarrow \lambda_{1,0}\ge0$, the estimated $\hat{\boldsymbol{\beta}}_{i}$ of (\ref{equ8}) converges to 
$$\arg \min_{\boldsymbol{\beta}_{i}} \frac{1}{2}\lVert \boldsymbol{A}_{i}-\boldsymbol{\beta}_{i}\boldsymbol{A}_{-i} \lVert_{2}^{2}+\frac{\sigma_0^2}{2} \lVert \boldsymbol{\beta}_{i} \lVert_{2}^{2} + \lambda_{1,0} \lVert \boldsymbol{\beta}_{i} \lVert_{1}$$ 
as $N_0$ goes to infinity.\\
(3). The estimated $\hat{\boldsymbol{\beta}}_{i}$ satisfies the LASSO Subspace Detection Property when $$0<\lim_{N_0\rightarrow\infty}\lambda_1(N_0) /N_0 = \lambda_{1,0}<\lVert \boldsymbol{A}_{i} \boldsymbol{A}'_{-i} \lVert_{\infty},$$ as $N_0$ goes to infinity.
}

The proof of the theorem is provided in Appendix A of the Supplementary Material. This theorem tells us when $\lambda_1(N_0) /N_0 \rightarrow 0$, the optimization is determined only by the first term and thus it has a closed-form solution. However, it does not guarantee the solution is sparse. On the other hand, if we select a sequence for $\lambda_1(N_0)$ satisfying $\lambda_1(N_0) /N_0 \rightarrow \lambda_{1,0}>0$, $\hat{\boldsymbol{\beta}}_{i}$ converges to the estimator of an elastic net regularized regression problem, and the sparsity property can be guaranteed. Besides, as long as $0<\lambda_{1,0}<\lVert \boldsymbol{A}_{i} \boldsymbol{A}'_{-i} \lVert_{\infty}$, the estimated representation coefficient $\hat{\boldsymbol{\beta}}_{i}$ satisfies the LASSO Subspace Detection Property, i.e., the time series is only represented by those from the same subspace.  
To better achieve the sparse structure of the streaming data and guarantee the LASSO Subspace Detection Property, we set $\lambda_1(N_0)=\lambda_{1,0} N_0,\ \lambda_{1,0}>0$ in this paper, i.e., $\lambda_1$ linearly increases with the segment length.
\\

For the change detection problem, let $\gamma_{c}^{0}={\tau_{c}^{0}}/{N}$ for $c=1, \ldots, C,\ \boldsymbol{\gamma}^{0}=\left(\gamma_{1}^{0}, \gamma_{2}^{0}, \ldots, \gamma_{C}^{0}\right)$ where the superscript 0 refers to the true value. Similarly, define $\gamma_{c}={\tau_{c}}/{N},\ \boldsymbol{\gamma}=\left(\gamma_{1}, \gamma_{2},\ldots, \gamma_{C}\right)$.  Define $\delta^{(c)}=\tau_c^0-\tau_{c-1}^0$ as the length of one segment. Note that $\boldsymbol{\gamma}^{0}$ is set to be a constant vector as $N$ goes to infinity.

\noindent
{\bf Theorem 2.}\\
{\it Under (A1-A2), for any time series $\boldsymbol{Y}$, there exists a constant $\lambda_{2,0}>0$, when $\lambda_2>\lambda_{2,0}$ and as $N$ goes to infinity with a fixed $\boldsymbol{\gamma}^0$, \\
(1). The optimal solution of optimization problem (\ref{equ7}) satisfies,
$$\rho\left(\boldsymbol{\gamma}^0,\ \hat{\boldsymbol{\gamma}}\right) \stackrel{p}{\longrightarrow} 0,$$
$$\rho\left(\hat{\boldsymbol{\gamma}},\ \boldsymbol{\gamma}^0\right) \stackrel{p}{\longrightarrow} 0,$$
where $\rho(\boldsymbol{a},\boldsymbol{b})=\max_i \left( \min_j |a_i-b_j| \right)$ is a function representing the difference of $\boldsymbol{a}$ and $\boldsymbol{b}$.\\
(3). If the number of change-points $\hat{C}$ is correctly estimated with properly specified $\lambda_2$, then
$$\hat{\gamma_{c}} \stackrel{p}{\longrightarrow} \gamma_{c}^{0},\text{ for } c=1,\ldots,C,\ $$
and the estimated coefficient matrices $\hat{\boldsymbol{B}}^{(c)},\ c=1,\ldots,C+1$ satisfy the asymptotic properties of Theorem 1.
}

The proof of Theorem 2 is provided in Appendix B of the Supplementary Material. Theorem 2(1) shows that for any true change-point, there exists a detected change-point from the optimization problem (7), such that the relative locations of the two converge in probability to each other. Similarly, for each detected change-point, there exists a true change-point, such that their relative locations converge to each other. In other words, although there may appear more than one detected change-point around each true change-point if $\lambda_2$ is not large enough, the relative locations of these detected change-points will converge to that of the true one. Theorem 2(2) tells us that if the number of change-point is correctly estimated with properly specified $\lambda_2$ in the optimization problem (7), then the relative location of each detected change-point will converge to that of the true change-point, and the estimated coefficient matrix of each segment will satisfy the asymptotic properties stated in Theorem 1. We have to mention that although Theorem 1 and 2 are offline properties, they still provide us insights into the online detection process, e.g., with observations being collected continuously after a change-point, the detection becomes more and more accurate.

\section{Online Optimization via PELT Algorithm}
\label{sec4}
 
In this section, we show the specific approach to solve the optimization problem (\ref{equ7}). We first introduce the PELT algorithm, which is an efficient sequential optimization algorithm. Then, we discuss how to determine the penalty coefficients and proper parameters for the implementation of PELT algorithm in DSSL model.

\subsection{The PELT based Optimization Algorithm}
Various algorithms have been proposed to solve the optimization problem of the following form for multiple change-points models:
\begin{equation}\min _{m, \tau_{1}, \ldots, \tau_{m}} \sum_{c=1}^{m+1}\left\{\operatorname{Cost}\left(\mathbf{Y}_{\tau_{c-1}+1: \tau_{c}}\right)\right\}+f(m),\end{equation}
where $\operatorname{Cost}(\cdot)$ is a cost function for a segment, $m$ is the number of change-points and $f(m)$, e.g., $f(m)=\beta (m+1)$, is a penalty term to guard against overfitting. Binary Segmentation (BS) algorithm proposed by \citet{scott1974cluster} is one of the most established search method with an $\mathcal{O}(n \log n)$ computational cost for $n$ samples. It begins by initially applying the single change-point method to the entire data set. The data set is split into two segments and the single change-point detection method is carried out again for these two segments independently. This procedure is repeated until no further change-points are detected. The advantage of the BS method is that it is computationally efficient. But it does not guarantee to find the global optimal solution.

The optimal partitioning (OP) algorithm proposed by \citet{jackson2005algorithm} focuses on finding the latest change-point (LCP) at each time step. It relates the optimal value of the cost function to the cost for the optimal partition of the data prior to the latest change-point plus the cost for the segment from the latest change-point to the end of the data. Let $F(n)$ denote the optimal value of the objective function for data $\boldsymbol{Y}_{1: n}$ and $\boldsymbol{\tau}_{n}=\left\{\boldsymbol{\tau}: 0=\tau_{0}<\tau_{1}<\cdots<\tau_{m}<\tau_{m+1}=n\right\}$ be the set of all possible vectors of the change-points for this dataset. Set $f(m)=\beta(m+1)$ and $F(0)=-\beta$. Then, 
\begin{equation}\begin{aligned}
F(n) &=\min _{\boldsymbol{\tau} \in \boldsymbol{\tau}_{n}}\left\{\sum_{c=1}^{m+1}\left[\operatorname{Cost}\left(\boldsymbol{Y}_{\tau_{c-1}+1: \tau_{c}}\right)+\beta\right]\right\} \\
&=\min _{t \in\{0, \ldots, n-1\}}\left\{\min _{\boldsymbol{\tau} \in \boldsymbol{\tau}_{t}} \sum_{c=1}^{m}\left[\operatorname{Cost}\left(\boldsymbol{Y}_{\tau_{c-1}+1: \tau_{c}}\right)+\beta\right]+\operatorname{Cost}\left(\boldsymbol{Y}_{t+1: n}\right)+\beta\right\} \\
&=\min _{t \in\{0, \ldots, n-1\}}\left\{F(t)+\operatorname{Cost}\left(\boldsymbol{Y}_{t+1: n}\right)+\beta\right\}.
\end{aligned}\end{equation}

As $F(t)$ only needs to be calculated once and can be used repeatedly in the following steps, this recursion can be solved sequentially or dynamically for $n=1, \ldots, N$, and the computational cost is $\mathcal{O}\left(N^{2}\right)$. Although the OP algorithm can find the global optimal solution, it is still far from being computationally competitive with the BS method. To this end, \citet{killick2012optimal} proposed the PELT (Pruned Exact Linear Time) by adding a pruning step for the OP algorithm to reduce the computational cost  to $\mathcal{O}\left(N\right)$. In the OP algorithm, to solve $F(n)=\min _{t \in\{0, \ldots, n-1\}}\left\{F(t)+\operatorname{Cost}\left(\boldsymbol{Y}_{t+1: n}\right)+\beta\right\}$, we need to consider all time points prior to time $n$. But in the PELT algorithm, we remove the time points that can never be the optimal LCP. Specifically, we optimize $F(n)=\min _{t \in R(t)}\left\{F(t)+\operatorname{Cost}\left(\boldsymbol{Y}_{t+1: n}\right)+\beta\right\}$ where $R(n)$ is the set of all time points that could be the possible LCP in terms of optimality at time $n$. The following theorem \citep{killick2012optimal}  provides a simple condition under which such pruning can be performed.\\

\noindent
{\bf Theorem 3.} \citep{killick2012optimal}
{\it If there exists a constant K such that for all $s<n<T$,
\begin{equation}\operatorname{Cost}\left(\boldsymbol{Y}_{s+1: n}\right)+\operatorname{Cost}\left(\boldsymbol{Y}_{n+1: T}\right)+K \leq \operatorname{Cost}\left(\boldsymbol{Y}_{s+1: T}\right).\label{equ14}\end{equation}
Then if
\begin{equation}F(s)+\operatorname{Cost}\left(\boldsymbol{Y}_{s+1: n}\right)+K \geq F(n)\label{equ15}\end{equation}
holds, at a future time $T>n$, $s$ can never be the optimal last change-point prior to $T$.
} \\

The proof can be found in Section 5 of Supplementary Material in \citep{killick2012optimal}. This result states that if Equation (\ref{equ15}) holds, then for any $T>n$, the best segmentation with the LCP prior to $T$ being at $n$ will be better than any with the LCP at $s$. Note that there exists a proper constant $K$ satisfying Equation (\ref{equ14}) for almost all cost functions used in practice. For example, if the cost function is the negative log-likelihood, then the constant can be selected as $K=0$. Therefore, we have 
\begin{equation}
\label{R_update}
R(n+1)=\{n\} \cup \left\{\tau \mid \tau \in R(n) , F(\tau)+\operatorname{Cost}\left(\boldsymbol{Y}_{\tau+1: n}\right)+K<F(n)\right\}.
\end{equation}
Our pruning process makes the optimization process very efficient under mild conditions with approximately linear computational cost with $n$, or on average a constant computational cost at each time step, which is highly desirable for online change-point detection. 

In our optimization problem (\ref{equ7}), the cost function and the penalty term can be expressed as
\begin{equation}
\begin{array}{c}
\label{equ16}
\operatorname{Cost}\left(\boldsymbol{Y}_{\tau_{c-1}+1: \tau_{c}}\right)=\sum_{i=1}^{p}\left[\sum_{t=\tau_{c-1}+1}^{\tau_{c}} \frac{1}{2}\left(Y_{i t}-\sum_{j \neq i} \hat{\beta}_{i j}^{(c)} Y_{j t}\right)^{2}+\lambda_{1} \sum_{j \neq i}\left|\hat{\beta}_{i j}^{(c)} \right|\right],\\
 f(C)=\lambda_{2}(C+1), 
\end{array}
\end{equation}
where $\hat{\beta}_{i j}^{(c)}$ is the estimated coefficients via LASSO algorithm for the following optimization problem
\begin{equation}
\label{equ17}
\min _{\beta_{i j}^{(c)}, j \neq i}\left\{\sum_{t=\tau_{c-1}+1}^{\tau_{c}} \frac{1}{2}\left(Y_{i t}-\sum_{j \neq i} \beta_{i j}^{(c)} Y_{j t}\right)^{2}+\lambda_{1} \sum_{j \neq i}\left|\beta_{i j}^{(c)}\right|\right\}.
\end{equation}

The detailed PELT based algorithm to solve our optimization problem (\ref{equ7}) is shown in Algorithm \ref{alg1}. As mentioned before, under certain conditions, most importantly that the number of change-points is increasing linearly with $N$, the computational efficiency of PELT is $O(N)$. For calculating $\operatorname{Cost}\left(\cdot\right)$ in each time step of DSSL algorithm, there are $p$ LASSO optimization problems and the computational efficiency of each problem is about $O(p)$. Therefore, assuming that the requirements of PELT algorithm are also satisfied for DSSL algorithm, the computational efficiency of DSSL algorithm is about $O(Np^2)$. In addition, it is worth noting that the computational efficiency is also related with the sparsity of the coefficients. The higher the sparsity of the model, the lower the computational complexity.

\begin{algorithm}[htb] 
\label{alg1}
\caption{\small The PELT based sequential optimization algorithm for DSSL} 
\textbf{Input:} Data set $\boldsymbol{Y}$, cost function $\operatorname{Cost}$, penalty constant $\lambda_2$, constant $K$, constant $s_0$\\
\textbf{Initialize:} $N$: the length of data, $F(0)=-\lambda_{2},\ c p(0)=N U L L,\ R(1)=\{0\}$\\ 
\textbf{Iterate} for $n=1, \ldots, N$ 
\begin{enumerate} 
  \item Calculate $F(n)=\min _{\tau \in R(n)}\left\{F(\tau)+\operatorname{Cost}\left(\boldsymbol{Y}_{\tau+1: n}\right)+\lambda_{2}\right\}.$
  \item Let $\hat{\tau}=\underset{\tau \in R(n)}{\operatorname{argmin}}\left\{{F}(\tau)+\operatorname{Cost}\left(\boldsymbol{Y}_{\tau+1: n}\right)+\lambda_{2}\right\}$ and $\widehat{\boldsymbol{B}}_{n}=\underset{B}{\operatorname{argmin}}\left\{\operatorname{Cost}\left(\boldsymbol{Y}_{\hat{\tau}+1: n}\right)\right\}.$
  \item Update $c p(n)=\{c p(\hat{\tau}), \hat{\tau}\}$ and $\mathcal{B}(n)=\left(\mathcal{B}(\hat{t}), \widehat{\boldsymbol{B}}_{n}\right).$  
  \item  Update $R(n+1)=\{n\} \cup \left\{\tau \mid \tau \in R(n) , F(\tau)+\operatorname{Cost}\left(\boldsymbol{Y}_{\tau+1: n}\right)+K<F(n)\right\}.$
\end{enumerate} 
\textbf{End}
\label{alg1} 
\end{algorithm}

\subsection{Parameter Selection}
\label{sec42}




The selection of optimal penalty weights $\lambda_{1,0}$ and $\lambda_{2}$ is nontrivial yet important for online change-points detection. In this subsection, an efficient tuning method combining  Bayesian Information Criterions (BIC) and change-point detection accuracy are proposed for the PELT algorithm.

Cross validation methods and information criterion methods are two  widely used methods in model selection or parameter selection. An advantage of information criterion methods is that they have considerably less computational expense than CV methods \citep{kirkland2015lasso}. Classical criterions, such as AIC and BIC criteria, have been applied for many fields and have proven their efficiency and accuracy. For high-dimensional data with large sample size, we prefer BIC criterion for a sparser model:
$$\mathrm{BIC}=-2 \log (Likelihood)+D \log (N)=\frac{\sum err^2}{\sigma^2}+N \log(\sigma^2)+D \log (N)+constant,$$
where $err$ is the residual of regression and $D$ is the dimension of the model. Estimate the variance $\sigma^2$ as $\frac{1}{N}\sum err^2$ and we have
$$\mathrm{BIC}=N\log(\frac{1}{N}\sum err^2)+D \log (N)+constant.$$
Here we propose the following criterion to select $\lambda_{1,0}$ for our multiple change-points model based on the BIC criterion:
$$(N p) \log (\frac{1}{Np}\sum err^2)+\sum_c\widehat{d f}^{(c)} \log (\delta^{(c)}),$$
where $D=\widehat{d f}(\lambda)$
is estimated as the number of nonzero elements in regression coefficients. The first term favors complex models, while the second term is a penalty term balancing the bias-variance tradeoff.
 Then, the parameter tuning turns to an optimization problem of the BIC criterion based on some historical or simulated data with change-points labels. The above BIC criterion is first used to select $\lambda_{1,0}$ using a searching grid method.
 With $\lambda_{1,0}$ determined, we search for the optimal $\lambda_{2}$ to have both low false alarm and low false detection. We select $\lambda_{2}$ to minimize
$\sum_{t=1}^{N} \boldsymbol{1}_{\{\lvert\hat{lcp}(t)-lcp(t)\lvert>s\}},$
where $lcp(t)$ refers to the detected latest change-point of time t and $s$ refers to the bound of false alarm.


The parameter $K$ is very critical in the pruning of the LCPs to achieve an efficient computation. Based on Equation (\ref{equ16}) and (\ref{equ17}), we prove in Theorem 4 (the proof is given in Appendix C of the Supplementary Material) that $K=0$ satisfies (\ref{equ14}) for any time series, and $K<\lambda_2$ for almost all cases except the special case that all time points are change-points of (\ref{equ7}), which is negligible. Therefore, in practical applications, $K$ can be set to a larger constant in $(0,\lambda_2)$ with better pruning performance and little influence on the detection accuracy. In the case studies, we set $K$ to $K=\frac{2}{3}\lambda_2$ which yields very good performance.

\noindent
{\bf Theorem 4.} 
{\it For the cost function determined by Equation (\ref{equ16}) and (\ref{equ17}),\\
(1). $K=0$ satisfies (\ref{equ14}) for all $s<n<T$ and any time series $\boldsymbol{Y}$;\\
(2). $K$ cannot be larger than $\lambda_2$ unless all the time points are change-points. 
}

\section{Case Studies}
In this section, the proposed algorithm is evaluated through numerical experiments with synthetic and real gesture data. All experiments are done using Python on an i7-6800K 3.40GHz Intel processor with 16GB RAM. 
\label{sec5}
\subsection{Simulation Study}
 
In this subsection, we test our proposed algorithm on three simulation scenarios. The synthetic data $\boldsymbol{X}$ are generated in two ways, one following the assumed subspace model described in Section \ref{sec21} (Case I, II) and the other following the GGM model (Case III). The variance of the noise is set to $\sigma^2=0.05^2, 0.1^2 \text{ and } 0.2^2$ for each case. The experiment of each scenario is repeated 100 times. In all the scenarios, the two penalty parameters are tuned with three additionally simulated high-dimensional signals.

\begin{figure}[htb]
  \centering
  \includegraphics[width=\textwidth]{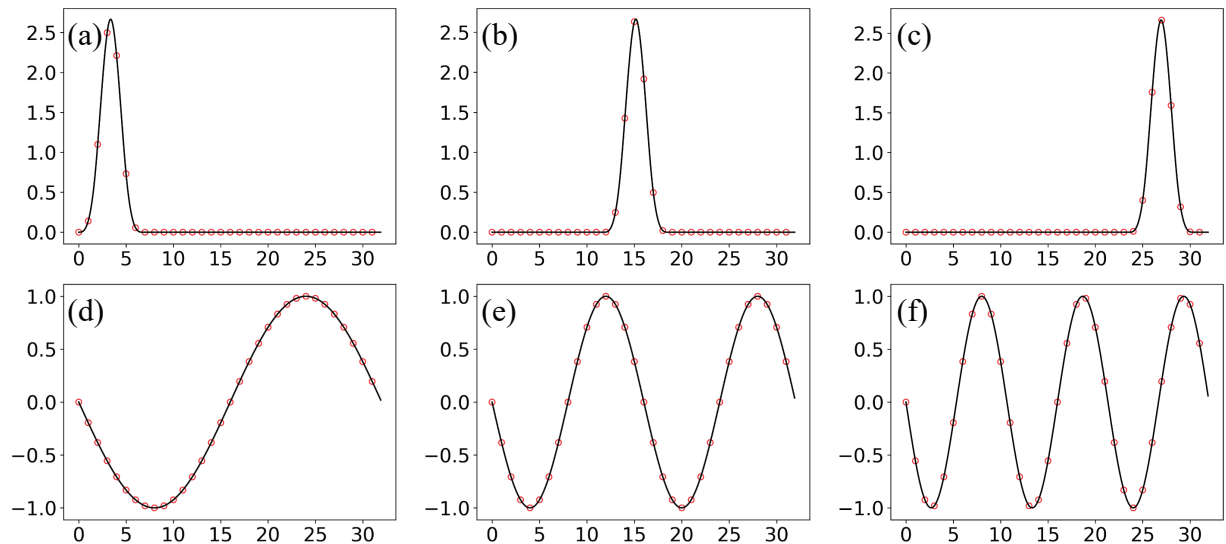} 
  \caption{Basis functions of the two subspaces. (a-c): B-spline basis functions; (d-f): Fourier basis functions.} 
  \label{fig2} 
\end{figure}

Case I. In this case, based on the subspace assumptions, two subspaces are simulated, where the first half of these time series are generated by linearly combining three B-spline basis functions, while the other half is generated by Fourier basis functions. The length of the time series is set to $N=128$ and suppose there are two change-points at $\tau_{1}=N / 4$ and $\tau_{2}=N / 2$ respectively. The dimension is set to $p=40$, i.e., 20 time series for each subspace. These basis functions are shown in Figure \ref{fig2}. The coefficients $\boldsymbol{A}$ are randomly generated in [-0.5,0.5] for each segment.

%

Case II. In this case, we test our algorithm with higher dimension and more change-points. The dimension is set to $p=400$. The length of the time series is set to $N=320$, and suppose there are nine change-points at $\tau_{i}=i\times \frac{N}{10},\ i=1,\dots,9$.

Case III. In this case, the synthetic data are generated based on the GGM model where the covariance matrix or precision matrix changes. The data in each segment are generated from $N(0,\sigma^2 \boldsymbol{\Sigma}^{(c)}),\ c=1,2$, where $\boldsymbol{\Sigma}^{(c)}=diag(\boldsymbol{\boldsymbol{\Sigma}}^{(c)}_{11},\boldsymbol{\Sigma}^{(c)}_{22})$, $\boldsymbol{\Sigma}^{(c)}_{11}$ and $\boldsymbol{\Sigma}^{(c)}_{22}$ are $20\times20$ matrices randomly generated by Vine method \citep{lewandowski2009generating} and $\sigma$ is set to 1,2 and 4 in three cases. Other settings are the same as Case I.

\begin{figure}[htbp]
  \centering
  \includegraphics[width=\textwidth]{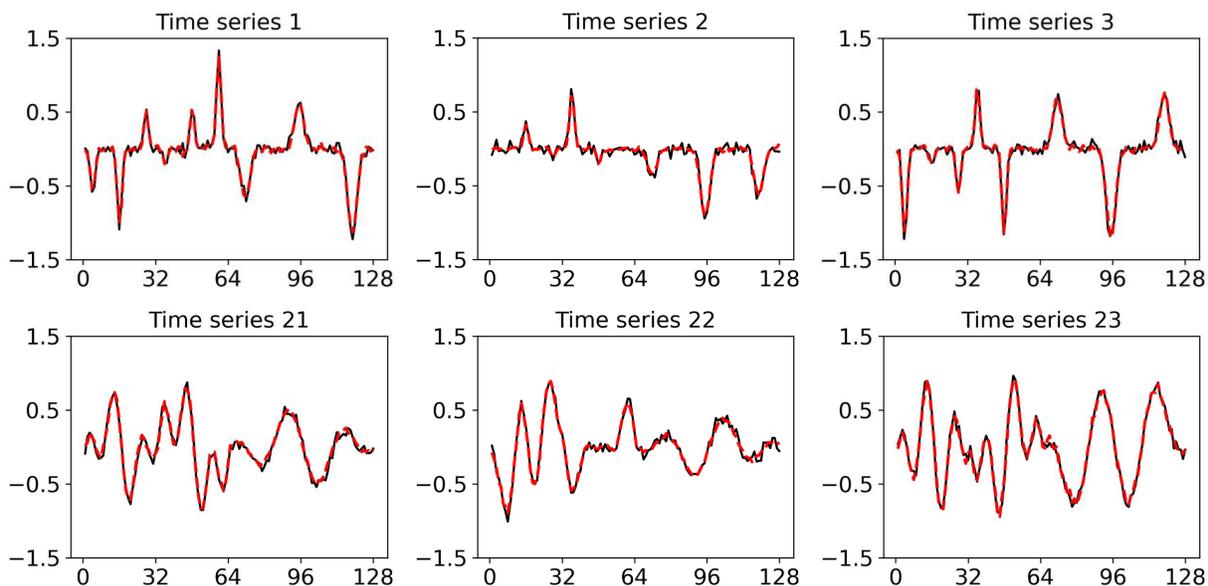} 
  \caption{ The original and fitted traces of 6 representative time series in one run (top: subspace I; bottom: subspace II). The black solid lines are the raw curves, and the red dashed lines are estimated values using the proposed method.}
  \label{fig3} 
\end{figure}

First, we provide the detection results of DSSL in detail for Case I when the variance of the noise is set to $\sigma^2=0.05^2$.  The two penalty coefficients are set to $\lambda_{1,0}=0.0028$ and $\lambda_2=2.2$. To avoid over-pruning when using $K=\frac{2}{3}\lambda_2$, we intentionally replace $n$ with $n-5,\ n-10,\ n-15$ in (\ref{R_update}) to get the candidate set of the LCP. Figure \ref{fig3} shows the 6 representative time series in one run and the fitted ones using the proposed method. Clearly, due to sufficient samples in each subspace, the self-expression is very accurate in all three segments. 

\begin{figure}[htbp]
  \centering
  \includegraphics[width=\textwidth]{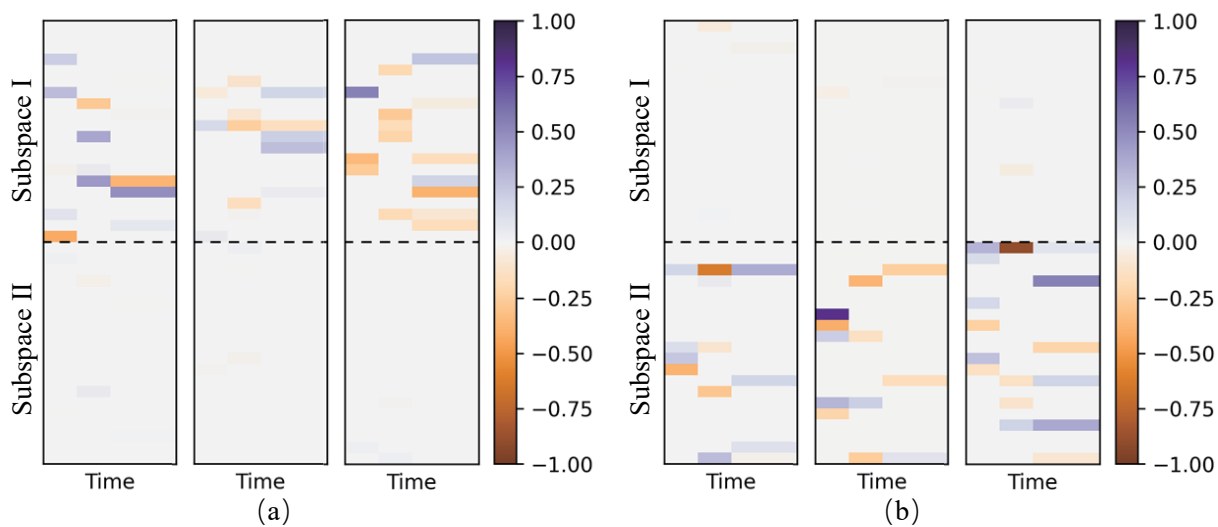} 
  \caption{ The estimated coefficients for the first three time series of each subspace ($\boldsymbol{Y}_{1},\boldsymbol{Y}_{2},\boldsymbol{Y}_{3}$ and $\boldsymbol{Y}_{21},\boldsymbol{Y}_{22},\boldsymbol{Y}_{23}$) as response variables: (a) subspace I and (b) subspace II.}
  \label{fig4} 
\end{figure}

Figure \ref{fig4} shows the estimated coefficients at the final time step for the first three time series of each subspace for illustration.  For example, the rows of the first panel represents the coefficients of $\boldsymbol{Y}_2, \ldots, \boldsymbol{Y}_{40}$ when $\boldsymbol{Y}_1$ is used as the response. These coefficients change at two change-points. Only the time series within the same subspace have nonzero coefficients, and each representation is sparse. These coefficients are constant within each segment.

\begin{figure}[htbp]
  \centering
  \includegraphics[width=\textwidth]{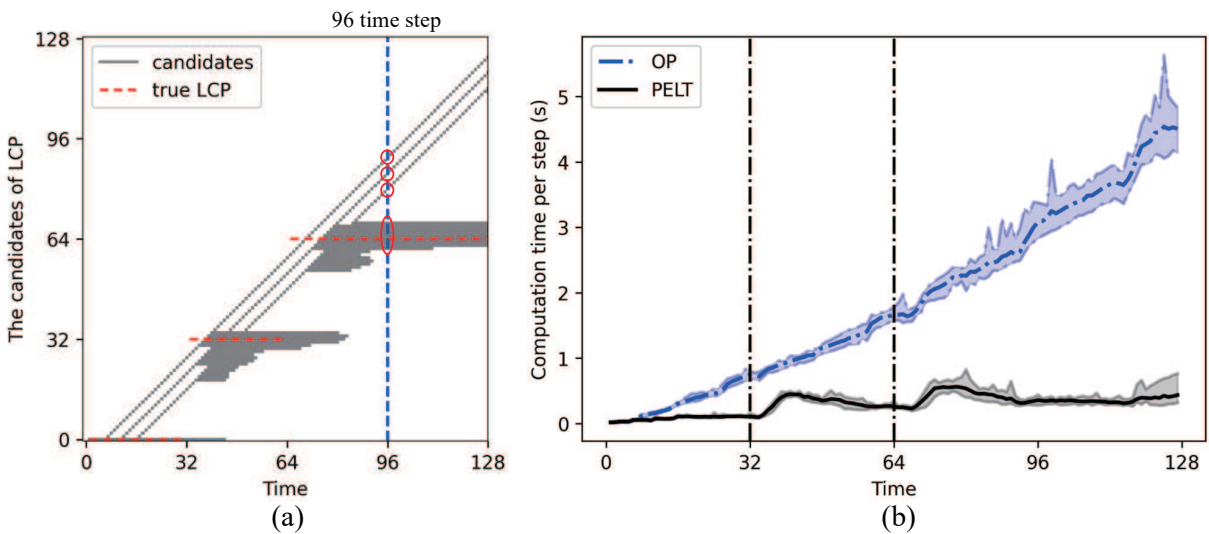} 
  \caption{Comparison of OP algorithm and PELT algorithm. (a) The selected candidates for the LCP at each time step and (b) average computation time and confidence interval per step.} 
  \label{fig5} 
\end{figure}

Figure \ref{fig5}(a) shows the selected candidates for the LCP at each time step in one run. Only a very small number of points including the true LCP are selected as the candidates, which can effectively control the computational cost at each step.  For instance, at 96 time step, the PELT algorithm only selects $\{91, 86, 81, 69, 68, 67, 66, 65, 64, 63, 62, 61\}$ as LCP candidates. Figure \ref{fig5}(b) shows the average computation time and confidence interval per step, where the PELT algorithm can save more than 90\% of the computation time than the OP algorithm.

\begin{figure}[htbp]
  \centering
  \includegraphics[width=0.7\textwidth]{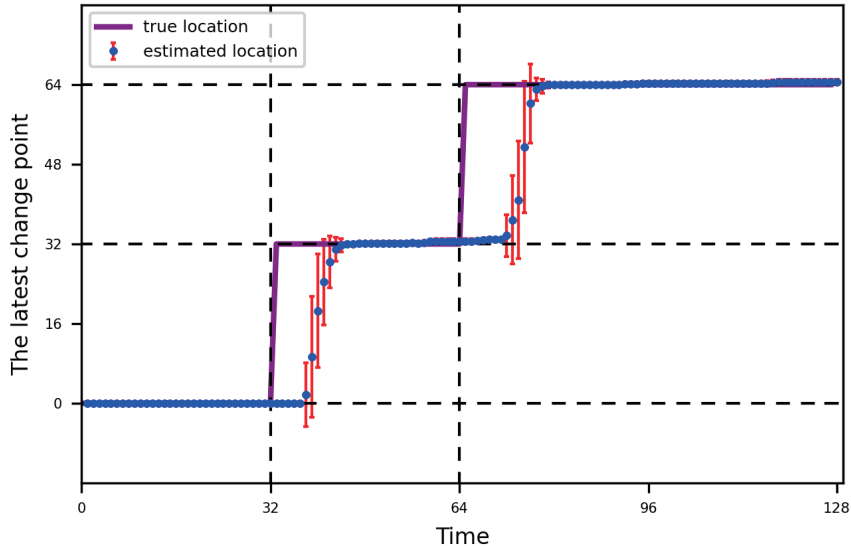} 
  \caption{The online detected latest change-point.} 
  \label{fig6} 
\end{figure}

Figure \ref{fig6} shows the mean and confidence interval of the detected LCP at each time step. It can be clearly seen that in most of the cases, the change can be timely detected with only about 10 time steps after the change occurs. Besides, as the observations since the latest change accumulates, the detected location of the LCP becomes more and more accurate. Note that in each replication, the detected LCP often abruptly jumps to the true values shortly after the change occurs. As the detection delay may vary from run to run, the change of the mean detection is not as abrupt as a single run.

In order to further demonstrate the superiority and effectiveness of our model, we compare our DSSL model with DFSL model \citep{zhang2020dynamic},  VCVS model \citep{kolar2012estimating} and GFGL model \citep{gibberd2017regularized}, which are state-of-the-art offline methods. Three offline and two online metrics are used for performance evaluation and comparison. The offline metrics include the precision, recall, and running time. The precision and recall are proposed by \citet{peel2015detecting} to evaluate offline multiple change-point detection algorithms and they are defined as follows:
$$
\begin{aligned}
Precision = \frac{1}{n_{e}} \sum_{i}^{n_{e}}\delta\left(\inf_j \left|\hat{\tau}^{(i)}-\tau^{(j)}\right| \le s  \right),\\
Recall = \frac{1}{n_{r}} \sum_{j}^{n_{r}}\delta\left(\inf_i \left|\hat{\tau}^{(i)}-\tau^{(j)}\right| \le s  \right),
\end{aligned}
$$
where $n_{e}$ and $n_{r}$ are the estimated number of change-points and the real number of change-points respectively, $\delta$ is the indicator function, and $s$ is the detection error bound. The precision is then the proportion of estimated change-points that timely detect true change-points. Similarly, recall is the proportion of true change-points that are timely detected. Here we set $s=5$ time steps in the comparison. The two online metrics, i.e., detection delay and running time per step, are used to evaluate the detection timeliness in online applications.

All experiments are replicated 100 times, and the results are shown in Table \ref{tab1}. For precision and recall metrics, the proposed DSSL is superior to other approaches in Case I and Case II. While in Case III, other approaches perform better than DSSL approach. The results show that our DSSL performs perfectly if the data satisfies the SSC assumptions. However, it is inferior to the GGM related methods when only the covariance matrix or precision matrix changes. Since the data are generated based on a different pattern, the BIC criterion may not be the optimal choice in such scenarios. DSSL performs better in Case II than Case I because the change is relatively more significant.

\begin{table}[tb]
\centering\scriptsize
\caption{The detection results of three simulation cases.}
\vspace{10pt}
\label{tab1}
\begin{tabular}{c|c|ccc|ccc|ccc}
\hline
 &  & \multicolumn{3}{c|}{Case I} & \multicolumn{3}{c|}{Case II} & \multicolumn{3}{c}{Case III} \\ \cline{2-11} 
\multirow{-2}{*}{} & sigma & \multicolumn{1}{c|}{0.05} & \multicolumn{1}{c|}{0.1} & 0.2 & \multicolumn{1}{c|}{0.05} & \multicolumn{1}{c|}{0.1} & 0.2 & \multicolumn{1}{c|}{1} & \multicolumn{1}{c|}{2} & 4 \\ \hline
 & DSSL & \multicolumn{1}{c|}{\textbf{100.00}} & \multicolumn{1}{c|}{\textbf{100.00}} & \textbf{89.00} & \multicolumn{1}{c|}{\textbf{100.00}} & \multicolumn{1}{c|}{\textbf{100.00}} & \textbf{100.00} & \multicolumn{1}{c|}{{93.50}} & \multicolumn{1}{c|}{{  91.17}} & { 92.33 } \\ \cline{2-11} 
 & DFSL & \multicolumn{1}{c|}{54.60} & \multicolumn{1}{c|}{56.72} & 61.24 & \multicolumn{1}{c|}{/} & \multicolumn{1}{c|}{/} & / & \multicolumn{1}{c|}{\textbf{100.00}} & \multicolumn{1}{c|}{\textbf{100.00}} & 93.93 \\ \cline{2-11} 
 & VCVS & \multicolumn{1}{c|}{47.87} & \multicolumn{1}{c|}{45.50} & 44.85 & \multicolumn{1}{c|}{/} & \multicolumn{1}{c|}{/} & / & \multicolumn{1}{c|}{94.98} & \multicolumn{1}{c|}{\textbf{100.00}} & \textbf{99.83} \\ \cline{2-11} 
\multirow{-4}{*}{Precision(\%)} & GFGL & \multicolumn{1}{c|}{34.76} & \multicolumn{1}{c|}{24.19} & 23.80 & \multicolumn{1}{c|}{88.81} & \multicolumn{1}{c|}{93.78} & 82.92 & \multicolumn{1}{c|}{{ 98.42}} & \multicolumn{1}{c|}{{ 96.85}} & { 96.58} \\ \hline
 & DSSL & \multicolumn{1}{c|}{\textbf{100.00}} & \multicolumn{1}{c|}{\textbf{100.00}} & 89.00 & \multicolumn{1}{c|}{\textbf{100.00}} & \multicolumn{1}{c|}{\textbf{100.00}} & \textbf{100.00} & \multicolumn{1}{c|}{{ 93.00 }} & \multicolumn{1}{c|}{{  92.00 }} & {  91.50 } \\ \cline{2-11} 
 & DFSL & \multicolumn{1}{c|}{81.00} & \multicolumn{1}{c|}{86.00} & 86.00 & \multicolumn{1}{c|}{/} & \multicolumn{1}{c|}{/} & / & \multicolumn{1}{c|}{\textbf{100.00}} & \multicolumn{1}{c|}{\textbf{100.00}} & \textbf{100.00} \\ \cline{2-11} 
 & VCVS & \multicolumn{1}{c|}{66.00} & \multicolumn{1}{c|}{65.00} & 74.50 & \multicolumn{1}{c|}{/} & \multicolumn{1}{c|}{/} & / & \multicolumn{1}{c|}{93.50} & \multicolumn{1}{c|}{99.50} & \textbf{100.00} \\ \cline{2-11} 
\multirow{-4}{*}{Recall(\%)} & GFGL & \multicolumn{1}{c|}{87.50} & \multicolumn{1}{c|}{99.00} & \textbf{99.50} & \multicolumn{1}{c|}{90.00} & \multicolumn{1}{c|}{96.67} & 90.00 & \multicolumn{1}{c|}{\textbf{100.00}} & \multicolumn{1}{c|}{\textbf{100.00}} & \textbf{100.00} \\ \hline
 & DSSL & \multicolumn{1}{c|}{{  35.29} } & \multicolumn{1}{c|}{{  28.93 }} & { 31.66 } & \multicolumn{1}{c|}{{ 2312.9 }} & \multicolumn{1}{c|}{{ 2020.9 }} & { 1898.1 } & \multicolumn{1}{c|}{{  1834.5 }} & \multicolumn{1}{c|}{{  1861.0 }} & {  1837.9 } \\ \cline{2-11} 
 & DFSL & \multicolumn{1}{c|}{199.04} & \multicolumn{1}{c|}{192.93} & 185.86 & \multicolumn{1}{c|}{/} & \multicolumn{1}{c|}{/} & / & \multicolumn{1}{c|}{269.43} & \multicolumn{1}{c|}{275.76} & 278.46 \\ \cline{2-11} 
 & VCVS & \multicolumn{1}{c|}{98.22} & \multicolumn{1}{c|}{91.94} & 83.41 & \multicolumn{1}{c|}{/} & \multicolumn{1}{c|}{/} & / & \multicolumn{1}{c|}{80.66} & \multicolumn{1}{c|}{108.46} & 149.83 \\ \cline{2-11} 
\multirow{-4}{*}{Running time(s)} & GFGL & \multicolumn{1}{c|}{\textbf{1.54}} & \multicolumn{1}{c|}{\textbf{3.70}} & \textbf{7.35} & \multicolumn{1}{c|}{\textbf{284.05}} & \multicolumn{1}{c|}{\textbf{296.72}} & \textbf{320.54} & \multicolumn{1}{c|}{\textbf{26.40}} & \multicolumn{1}{c|}{\textbf{26.34}} & \textbf{26.48} \\ \hline
\begin{tabular}[c]{@{}c@{}}Running time\\ per step(s)\end{tabular} & DSSL & \multicolumn{1}{c|}{{  \textbf{0.28} }} & \multicolumn{1}{c|}{{  \textbf{0.23 } }} & {  \textbf{0.25} } & \multicolumn{1}{c|}{{  \textbf{ 7.22} }} & \multicolumn{1}{c|}{{  \textbf{6.31} }} & {  \textbf{ 5.93} } & \multicolumn{1}{c|}{{ \textbf{14.32}  }} & \multicolumn{1}{c|}{{  \textbf{14.53} }} & {  \textbf{14.35} } \\ \hline
Detection delay & DSSL & \multicolumn{1}{c|}{10.17} & \multicolumn{1}{c|}{12.64} & 15.57 & \multicolumn{1}{c|}{10.52} & \multicolumn{1}{c|}{13.53} & 11.83 & \multicolumn{1}{c|}{{  16.82 }} & \multicolumn{1}{c|}{{  17.42}} & { 17.22 } \\ \hline
\end{tabular}
\end{table}

In terms of the running time, GFGL approach is the fastest algorithm and our DSSL approach is slightly slower in Case I and II. However, GFGL is an offline method, and if it is applied online, the whole optimization process has to be repeated at each time step, leading to a much higher overall cost. Note that, in Case II, DFSL approach and VCVS approach are unable to detect change-points due to either insufficient memory or extremely long computational time. In Case III, the DSSL approach has a longer running time because the optimization for LASSO problem is much more time-consuming when data are generated by the GGM model. The maximum number of iterations for solving LASSO is set to 1000 in Case I and II while it is set to 10000 in Case III. The running time can be reduced to one tenth by setting the maximum number of iterations to 1000, at a cost of 10 percent decreasing for precision and recall metrics. As for the detection delay, we only show the results for DSSL, as all other methods are offline. Similarly, DSSL approach performs better in Case II than Case I.

In addition, as the variance of noise increases, the change-points are more difficult to detect with lower precision, lower recall and larger detection delay generally. However, the performance of the proposed DSSL approach does not change a lot in Case III. The reason is that $\sigma^2$ in Case III only affects the variation of the data. The correlations among the time series remain the same.

\subsection{Human Motion Tracking}
\begin{figure}[htb]
  \centering
  \includegraphics[width=0.7\textwidth]{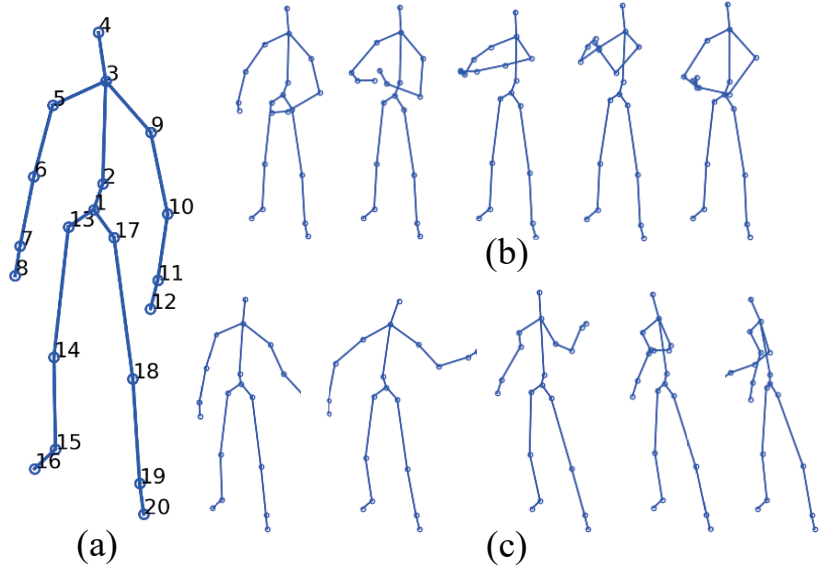} 
  \caption{(a) The position of the 20 joints; (b) five snapshots of the shoot gesture; (c) five snapshots of the throw gesture.} 
  \label{fig7} 
\end{figure}

In this subsection, we apply the proposed DSSL to the MSRC-12 Gesture Dataset for gesture tracking \citep{fothergill2012instructing}. This dataset consists of sequences of human skeletal body part movements and the associated gesture that needs to be recognized by the system. Each sample of the sequences contains 60 variables, which are the three dimensional coordinates of 20 human joints. The body pose is captured at a sample rate of 30Hz with $\pm 2 \mathrm{cm}$ accuracy in joint positions. In the MSRC-12 Gesture Dataset, there are 30 subjects and they perform 12 gestures each for ten times. The position of these 20 joints and some snapshots of shoot gesture and throw gesture are shown in Figure \ref{fig7}.

\begin{figure}[htb]
  \centering
  \includegraphics[width=\textwidth]{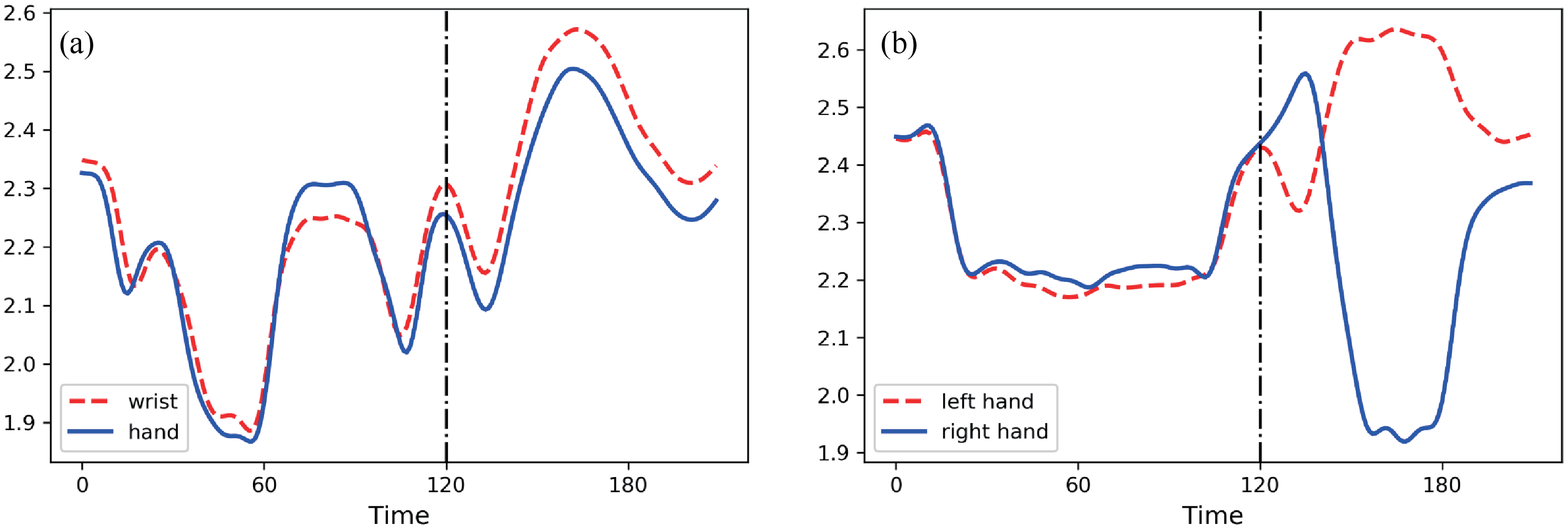} 
  \caption{The representative correlations between the joints: (a) the trace of two joints of the left arm and (b) the trace of the joints of the left and right hand.} 
  \label{fig8} 
\end{figure}

Motion segmentation is often a very critical step for gesture recognition. Here we combine the sequences of two gestures of the same subject together to demonstrate the effectiveness of the proposed method. Specifically, we choose ‘shoot’ and ‘throw’ as the two gestures. In the first segment, the subject stretches his arms out in front of him, holds a pistol in both hands, makes a recoil movement and then returns to the original position. In the second segment, the subject uses his right arm to make an overarm throwing movement and then returns to the original position. For the sake of simplicity, we convert coordinates of the same joint into one distance variable, which represents the distance of the joint from the reference point. 

Clearly, some joints share similar trajectories with each other, as these joints move in similar ways, such as the joints on the same arm or the same leg (Figure \ref{fig8}(a)). Some joints have totally different trajectories because they have no correlations. With this regard, we can infer that these joints lie in some subspaces and they can be naturally clustered into different groups. Besides, some joints share similar trajectories in the first segment but have different trajectories in the second segment, such as the joints of the two hands in the combined data. As shown in Figure \ref{fig8}(b), in the first segment, they increase or decrease synchronously, while in the second segment, they increase or decrease in the opposite direction. Therefore, we can apply the proposed method to detect when the gesture changes. We delete the data of the left and right wrist because these data are generally the same with the data of the left and right hand. So in total $p=18$ variables are considered.  We only use one historical streaming data to tune the two penalty parameters.
\\

\begin{figure}[htbp]
  \centering
  \includegraphics[width=\textwidth]{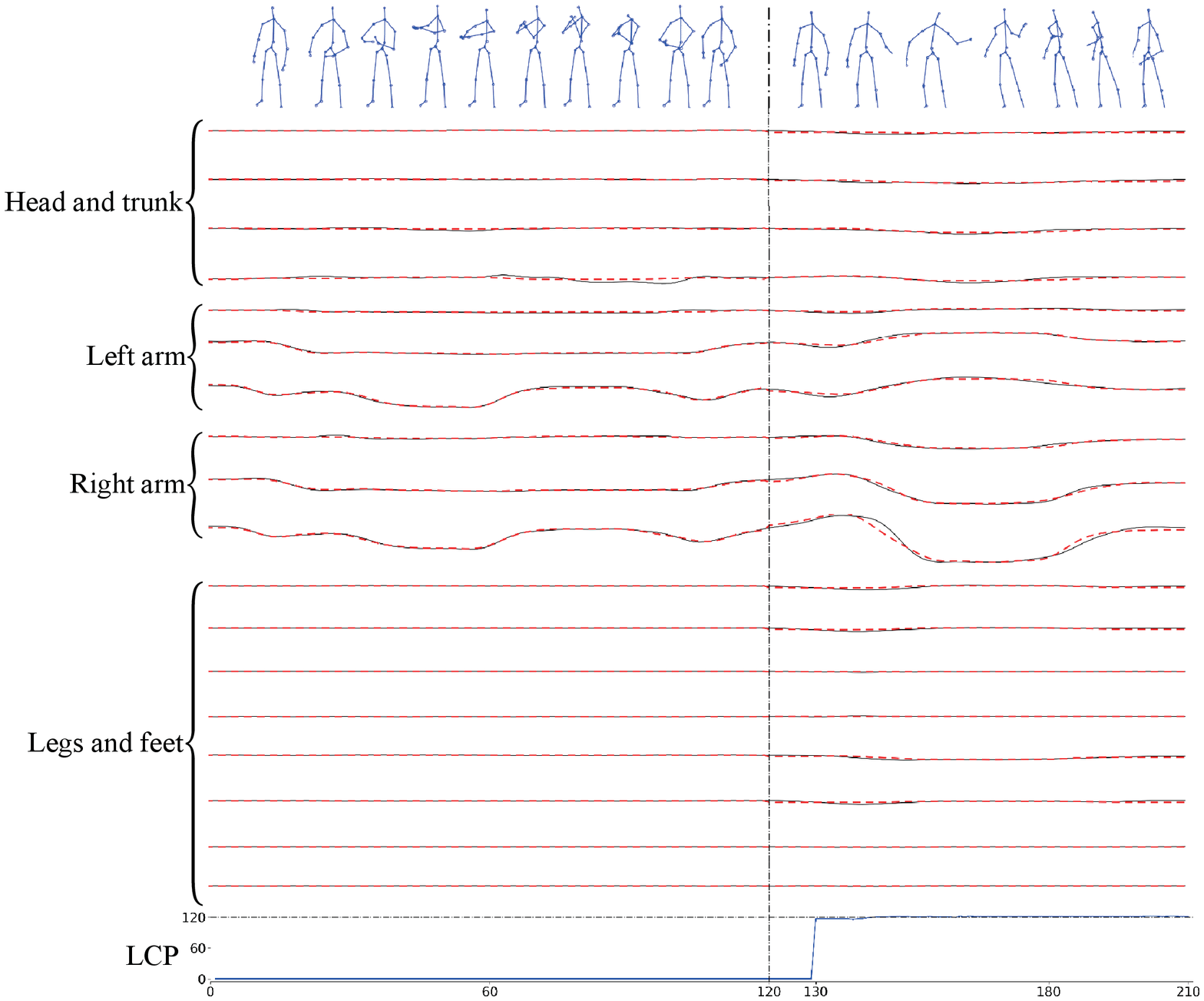} 
  \caption{The original and fitted traces of the 18 variables and the sequentially detected LCP. The black solid lines are the original curves, and the red dashed lines are the estimated ones. The vertical dashed line denote the true change-point.} 
  \label{fig9} 
\end{figure}

\begin{figure}[htbp]
  \centering
  \includegraphics[width=0.5\textwidth]{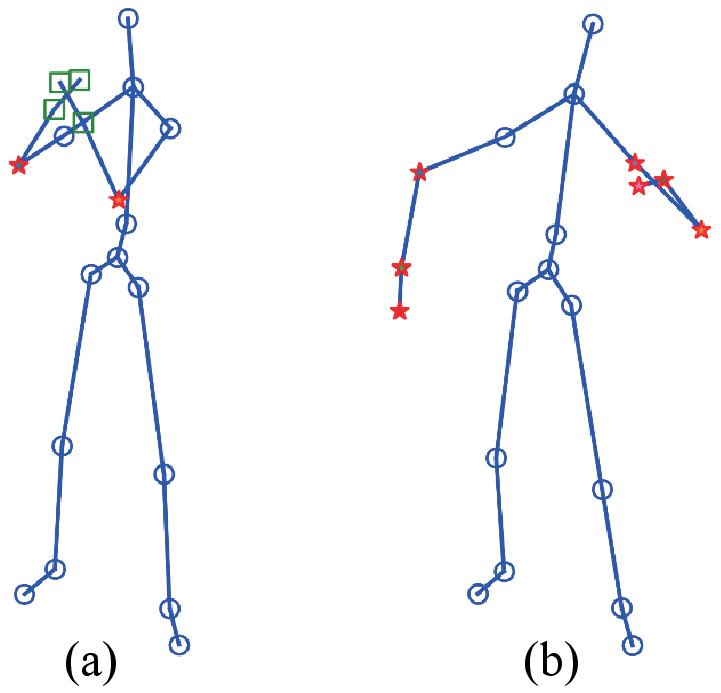} 
  \caption{The clustering of the joints: (a) the first segment and (b) the second segment. The joints of the same color and shape are grouped into one cluster.} 
  \label{fig10} 
\end{figure}

Figure \ref{fig9} shows the original and fitted variables and the sequentially detected LCPs. The change of the gesture is timely detected only after about 10 time steps. The positions of the joints of head, trunk, legs and feet do not change much in the whole motion sequence. The joints of left arm and right arm play a key role to detect the change-point. The subspaces for each gesture are further identified via the clustering of the coefficient matrix, which are shown in Figure \ref{fig10}. This result is consistent with the actual movement of each joint in these two gestures. 

The detection performance of the DSSL approach and other baselines is compared in Table \ref{tab2}. The detection error bound $s$ is set to 10 in the calculation of precision ratio and recall ratio.  Similar to the simulation study, the proposed DSSL approach outperforms other benchmarks in the detection accuracy with less running time per step. For all methods, the recall is high while the precision is relatively low, indicating that these two methods sometimes detect other change-points besides the one connecting the two gestures. This is understandable because there may exist some changes in one single gesture. For example, for gesture ‘throw’, the motion can be further divided to three segments due to slight correlation change, i.e., raising a hand, throwing something and putting down this hand. These changes may also be detected, though not often. 

\begin{table}[htbp]
\begin{center}
\caption{The detection results in human motion tracking.}
\label{tab2}
    \begin{tabular}{c|c|c|c|c}
    \hline
                               & DSSL & DFSL &VCVS &GFGL        \\ \hline
    Precision(\%)              & \textbf{81.82} & 50.00 &46.67 &20.00       \\ \hline
    Recall(\%)                 & \textbf{90.00} & 80.00 &60.00 &40.00      \\ \hline
    Total computation time(second)  & 28.7  & 367.8 & 3.06 & \textbf{2.30}  \\ \hline
    Detection delay(step)          & \textbf{7.6}  & $\backslash$ & $\backslash$ & $\backslash$   \\ \hline
    Computation time per step(second) &\textbf{0.14} & $\backslash$ & $\backslash$ & $\backslash$  \\ \hline
    \end{tabular}
\end{center}
\end{table}

\section{Discussion and Conclusions}
\label{sec6}
In this paper, we proposed a dynamic sparse subspace learning (DSSL) approach for online structural change-point detection of high-dimensional streaming data. Specifically, it is assumed that the high-dimensional data lie in multiple low-dimensional subspaces and the subspace structures may abruptly change over time. Only the variables from the same subspace correlate with each other and each variables can be sparsely represented by others. 

Based on the self-expressive property, we proposed a novel multiple structural change-point model with two penalty terms in the loss function to encourage sparse representation and avoid excessive change-points.  The asymptotic consistency of the estimators was further established. A PELT based algorithm was proposed for online optimization and change-point detection. Based on some historical data, the penalty coefficients in our model could be properly selected by a tuning method  combining a BIC criterion and detection accuracy. The effectiveness of the proposed method was demonstrated on synthetic data and gesture data for motion tracking.

There are several issues that are worthy of further investigation. Firstly, the proposed method assumes that the measurement noises of all variables are independent and identically distributed. However, the variance heterogeneity, cross-correlation and even auto-correlation may exist in practice. In addition, all the subspaces are assumed to be linear manifolds in the current work. To make it more flexible, nonlinear manifolds can be considered by using nonlinear regressions, such as kernel based methods. Last but not least, when there are not sufficient samples in the subspace, the self-expressive assumption may not hold. How to learn the subspace sequentially with insufficient samples is a challenging problem that needs to be solved.

\section*{Acknowledgments}
The authors would like to thank the editor, associate editor, and anonymous reviewers for many constructive comments which greatly improved the paper.

\section*{Supplementary Materials}
Additional proofs: Detailed proofs are included as a pdf file.\\
Python code: The zip file contains Python codes of all case studies.

\section*{Funding}
This work is supported in part by the NSFC grant NSFC-72171003, NSFC-71932006 and NSFC-51875003.



\bibliographystyle{chicago}
\bibliography{reference}

\newpage
\begin{thebibliography}{}

\bibitem[\protect\citeauthoryear{Aminikhanghahi and Cook}{Aminikhanghahi and
  Cook}{2017}]{aminikhanghahi2017survey}
Aminikhanghahi, S. and D.~J. Cook (2017).
\newblock A survey of methods for time series change point detection.
\newblock {\em Knowledge and information systems\/}~{\em 51\/}(2), 339--367.

\bibitem[\protect\citeauthoryear{Beck and Teboulle}{Beck and
  Teboulle}{2009}]{beck2009fast}
Beck, A. and M.~Teboulle (2009).
\newblock A fast iterative shrinkage-thresholding algorithm for linear inverse
  problems.
\newblock {\em SIAM journal on imaging sciences\/}~{\em 2\/}(1), 183--202.

\bibitem[\protect\citeauthoryear{Bellman}{Bellman}{1966}]{bellman1966dynamic}
Bellman, R. (1966).
\newblock Dynamic programming.
\newblock {\em Science\/}~{\em 153\/}(3731), 34--37.

\bibitem[\protect\citeauthoryear{Dempster}{Dempster}{1972}]{dempster1972covariance}
Dempster, A.~P. (1972).
\newblock Covariance selection.
\newblock {\em Biometrics\/}, 157--175.

\bibitem[\protect\citeauthoryear{Dhulekar, Nambirajan, Oztan, and
  Yener}{Dhulekar et~al.}{2015}]{dhulekar2015seizure}
Dhulekar, N., S.~Nambirajan, B.~Oztan, and B.~Yener (2015).
\newblock Seizure prediction by graph mining, transfer learning, and
  transformation learning.
\newblock In {\em International Workshop on Machine Learning and Data Mining in
  Pattern Recognition}, pp.\  32--52. Springer.

\bibitem[\protect\citeauthoryear{Drton and Perlman}{Drton and
  Perlman}{2008}]{drton2008sinful}
Drton, M. and M.~D. Perlman (2008).
\newblock A sinful approach to gaussian graphical model selection.
\newblock {\em Journal of Statistical Planning and Inference\/}~{\em 138\/}(4),
  1179--1200.

\bibitem[\protect\citeauthoryear{Elhamifar and Vidal}{Elhamifar and
  Vidal}{2013}]{elhamifar2013sparse}
Elhamifar, E. and R.~Vidal (2013).
\newblock Sparse subspace clustering: Algorithm, theory, and applications.
\newblock {\em IEEE transactions on pattern analysis and machine
  intelligence\/}~{\em 35\/}(11), 2765--2781.

\bibitem[\protect\citeauthoryear{Fothergill, Mentis, Kohli, and
  Nowozin}{Fothergill et~al.}{2012}]{fothergill2012instructing}
Fothergill, S., H.~Mentis, P.~Kohli, and S.~Nowozin (2012).
\newblock Instructing people for training gestural interactive systems.
\newblock In {\em Proceedings of the SIGCHI Conference on Human Factors in
  Computing Systems}, pp.\  1737--1746.

\bibitem[\protect\citeauthoryear{Foygel and Drton}{Foygel and
  Drton}{2010}]{foygel2010extended}
Foygel, R. and M.~Drton (2010).
\newblock Extended bayesian information criteria for gaussian graphical models.
\newblock {\em arXiv preprint arXiv:1011.6640\/}.

\bibitem[\protect\citeauthoryear{Gibberd and Nelson}{Gibberd and
  Nelson}{2017}]{gibberd2017regularized}
Gibberd, A.~J. and J.~D. Nelson (2017).
\newblock Regularized estimation of piecewise constant gaussian graphical
  models: The group-fused graphical lasso.
\newblock {\em Journal of Computational and Graphical Statistics\/}~{\em
  26\/}(3), 623--634.

\bibitem[\protect\citeauthoryear{G{\'o}mez, Paynabar, and Pacella}{G{\'o}mez
  et~al.}{2020}]{gomez2020functional}
G{\'o}mez, A. M.~E., K.~Paynabar, and M.~Pacella (2020).
\newblock Functional directed graphical models and applications in root-cause
  analysis and diagnosis.
\newblock {\em Journal of Quality Technology\/}, 1--17.

\bibitem[\protect\citeauthoryear{Guo, Gao, and Li}{Guo
  et~al.}{2013}]{guo2013spatial}
Guo, Y., J.~Gao, and F.~Li (2013).
\newblock Spatial subspace clustering for hyperspectral data segmentation.
\newblock In {\em Conference of The Society of Digital Information and Wireless
  Communications (SDIWC)}, Volume~1, pp.\ ~3.

\bibitem[\protect\citeauthoryear{Haslbeck and Waldorp}{Haslbeck and
  Waldorp}{2015}]{haslbeck2015mgm}
Haslbeck, J. and L.~J. Waldorp (2015).
\newblock mgm: Estimating time-varying mixed graphical models in
  high-dimensional data.
\newblock {\em arXiv preprint arXiv:1510.06871\/}.

\bibitem[\protect\citeauthoryear{Jackson, Scargle, Barnes, Arabhi, Alt,
  Gioumousis, Gwin, Sangtrakulcharoen, Tan, and Tsai}{Jackson
  et~al.}{2005}]{jackson2005algorithm}
Jackson, B., J.~D. Scargle, D.~Barnes, S.~Arabhi, A.~Alt, P.~Gioumousis,
  E.~Gwin, P.~Sangtrakulcharoen, L.~Tan, and T.~T. Tsai (2005).
\newblock An algorithm for optimal partitioning of data on an interval.
\newblock {\em IEEE Signal Processing Letters\/}~{\em 12\/}(2), 105--108.

\bibitem[\protect\citeauthoryear{Jiao, Chen, and Gu}{Jiao
  et~al.}{2018}]{jiao2018subspace}
Jiao, Y., Y.~Chen, and Y.~Gu (2018).
\newblock Subspace change-point detection: A new model and solution.
\newblock {\em IEEE Journal of Selected Topics in Signal Processing\/}~{\em
  12\/}(6), 1224--1239.

\bibitem[\protect\citeauthoryear{Killick, Fearnhead, and Eckley}{Killick
  et~al.}{2012}]{killick2012optimal}
Killick, R., P.~Fearnhead, and I.~A. Eckley (2012).
\newblock Optimal detection of changepoints with a linear computational cost.
\newblock {\em Journal of the American Statistical Association\/}~{\em
  107\/}(500), 1590--1598.

\bibitem[\protect\citeauthoryear{Kirkland, Kanfer, and Millard}{Kirkland
  et~al.}{2015}]{kirkland2015lasso}
Kirkland, L.-A., F.~Kanfer, and S.~Millard (2015).
\newblock Lasso tuning parameter selection.
\newblock In {\em Annual Proceedings of the South African Statistical
  Association Conference}, Volume 2015, pp.\  49--56. South African Statistical
  Association (SASA).

\bibitem[\protect\citeauthoryear{Kitov, Turuntaev, Konovalov, Stepnov, and
  Pupatenko}{Kitov et~al.}{2019}]{kitov2019use}
Kitov, I., S.~Turuntaev, A.~Konovalov, A.~Stepnov, and V.~Pupatenko (2019).
\newblock Use of waveform cross correlation to reconstruct the aftershock
  sequence of the august 14, 2016, sakhalin earthquake.
\newblock {\em Seismic Instruments\/}~{\em 55\/}(5), 544--558.

\bibitem[\protect\citeauthoryear{Kolar and Xing}{Kolar and
  Xing}{2011}]{kolar2011time}
Kolar, M. and E.~Xing (2011).
\newblock On time varying undirected graphs.
\newblock In {\em Proceedings of the Fourteenth International Conference on
  Artificial Intelligence and Statistics}, pp.\  407--415. JMLR Workshop and
  Conference Proceedings.

\bibitem[\protect\citeauthoryear{Kolar and Xing}{Kolar and
  Xing}{2012}]{kolar2012estimating}
Kolar, M. and E.~P. Xing (2012).
\newblock Estimating networks with jumps.
\newblock {\em Electronic journal of statistics\/}~{\em 6}, 2069.

\bibitem[\protect\citeauthoryear{Lewandowski, Kurowicka, and Joe}{Lewandowski
  et~al.}{2009}]{lewandowski2009generating}
Lewandowski, D., D.~Kurowicka, and H.~Joe (2009).
\newblock Generating random correlation matrices based on vines and extended
  onion method.
\newblock {\em Journal of multivariate analysis\/}~{\em 100\/}(9), 1989--2001.

\bibitem[\protect\citeauthoryear{Li and Solea}{Li and
  Solea}{2018}]{li2018nonparametric}
Li, B. and E.~Solea (2018).
\newblock A nonparametric graphical model for functional data with application
  to brain networks based on fmri.
\newblock {\em Journal of the American Statistical Association\/}~{\em
  113\/}(524), 1637--1655.

\bibitem[\protect\citeauthoryear{Li and Vidal}{Li and
  Vidal}{2015}]{li2015structured}
Li, C.-G. and R.~Vidal (2015).
\newblock Structured sparse subspace clustering: A unified optimization
  framework.
\newblock In {\em Proceedings of the IEEE conference on computer vision and
  pattern recognition}, pp.\  277--286.

\bibitem[\protect\citeauthoryear{Li, You, and Vidal}{Li
  et~al.}{2017}]{li2017structured}
Li, C.-G., C.~You, and R.~Vidal (2017).
\newblock Structured sparse subspace clustering: A joint affinity learning and
  subspace clustering framework.
\newblock {\em IEEE Transactions on Image Processing\/}~{\em 26\/}(6),
  2988--3001.

\bibitem[\protect\citeauthoryear{Meinshausen and B{\"u}hlmann}{Meinshausen and
  B{\"u}hlmann}{2006}]{meinshausen2006high}
Meinshausen, N. and P.~B{\"u}hlmann (2006).
\newblock High-dimensional graphs and variable selection with the lasso.
\newblock {\em The annals of statistics\/}~{\em 34\/}(3), 1436--1462.

\bibitem[\protect\citeauthoryear{Nowinski}{Nowinski}{1981}]{nowinski1981applications}
Nowinski, J.~L. (1981).
\newblock Applications of functional analysis in engineering.
\newblock {\em International Journal of Engineering Science\/}~{\em 19\/}(11),
  1377--1390.

\bibitem[\protect\citeauthoryear{Parsons, Haque, and Liu}{Parsons
  et~al.}{2004}]{parsons2004subspace}
Parsons, L., E.~Haque, and H.~Liu (2004).
\newblock Subspace clustering for high dimensional data: a review.
\newblock {\em Acm sigkdd explorations newsletter\/}~{\em 6\/}(1), 90--105.

\bibitem[\protect\citeauthoryear{Peel and Clauset}{Peel and
  Clauset}{2015}]{peel2015detecting}
Peel, L. and A.~Clauset (2015).
\newblock Detecting change points in the large-scale structure of evolving
  networks.
\newblock In {\em Twenty-Ninth AAAI Conference on Artificial Intelligence}.

\bibitem[\protect\citeauthoryear{Qiao, Guo, and James}{Qiao
  et~al.}{2019}]{qiao2019functional}
Qiao, X., S.~Guo, and G.~M. James (2019).
\newblock Functional graphical models.
\newblock {\em Journal of the American Statistical Association\/}~{\em
  114\/}(525), 211--222.

\bibitem[\protect\citeauthoryear{Qiao, Qian, James, and Guo}{Qiao
  et~al.}{2020}]{qiao2020doubly}
Qiao, X., C.~Qian, G.~M. James, and S.~Guo (2020).
\newblock Doubly functional graphical models in high dimensions.
\newblock {\em Biometrika\/}~{\em 107\/}(2), 415--431.

\bibitem[\protect\citeauthoryear{Qiu, Han, Liu, and Caffo}{Qiu
  et~al.}{2016}]{qiu2016joint}
Qiu, H., F.~Han, H.~Liu, and B.~Caffo (2016).
\newblock Joint estimation of multiple graphical models from high dimensional
  time series.
\newblock {\em Journal of the Royal Statistical Society. Series B, Statistical
  Methodology\/}~{\em 78\/}(2), 487.

\bibitem[\protect\citeauthoryear{Ren, Sun, Zhang, and Zhou}{Ren
  et~al.}{2015}]{ren2015asymptotic}
Ren, Z., T.~Sun, C.-H. Zhang, and H.~H. Zhou (2015).
\newblock Asymptotic normality and optimalities in estimation of large gaussian
  graphical models.
\newblock {\em The Annals of Statistics\/}~{\em 43\/}(3), 991--1026.

\bibitem[\protect\citeauthoryear{Rothman, Bickel, Levina, and Zhu}{Rothman
  et~al.}{2008}]{rothman2008sparse}
Rothman, A.~J., P.~J. Bickel, E.~Levina, and J.~Zhu (2008).
\newblock Sparse permutation invariant covariance estimation.
\newblock {\em Electronic Journal of Statistics\/}~{\em 2}, 494--515.

\bibitem[\protect\citeauthoryear{Scott and Knott}{Scott and
  Knott}{1974}]{scott1974cluster}
Scott, A.~J. and M.~Knott (1974).
\newblock A cluster analysis method for grouping means in the analysis of
  variance.
\newblock {\em Biometrics\/}, 507--512.

\bibitem[\protect\citeauthoryear{Shi, Zhang, Sun, and Song}{Shi
  et~al.}{2019}]{shi2019fuzzy}
Shi, M., L.~Zhang, W.~Sun, and X.~Song (2019).
\newblock A fuzzy c-means algorithm guided by attribute correlations and its
  application in the big data analysis of tunnel boring machine.
\newblock {\em Knowledge-Based Systems\/}~{\em 182}, 104859.

\bibitem[\protect\citeauthoryear{Thatcher, North, and Biver}{Thatcher
  et~al.}{2005}]{thatcher2005eeg}
Thatcher, R.~W., D.~North, and C.~Biver (2005).
\newblock Eeg and intelligence: relations between eeg coherence, eeg phase
  delay and power.
\newblock {\em Clinical neurophysiology\/}~{\em 116\/}(9), 2129--2141.

\bibitem[\protect\citeauthoryear{Tibshirani, Saunders, Rosset, Zhu, and
  Knight}{Tibshirani et~al.}{2005}]{tibshirani2005sparsity}
Tibshirani, R., M.~Saunders, S.~Rosset, J.~Zhu, and K.~Knight (2005).
\newblock Sparsity and smoothness via the fused lasso.
\newblock {\em Journal of the Royal Statistical Society: Series B (Statistical
  Methodology)\/}~{\em 67\/}(1), 91--108.

\bibitem[\protect\citeauthoryear{Tierney, Gao, and Guo}{Tierney
  et~al.}{2014}]{tierney2014subspace}
Tierney, S., J.~Gao, and Y.~Guo (2014).
\newblock Subspace clustering for sequential data.
\newblock In {\em Proceedings of the IEEE conference on computer vision and
  pattern recognition}, pp.\  1019--1026.

\bibitem[\protect\citeauthoryear{Wang and Xu}{Wang and
  Xu}{2013}]{wang2013noisy}
Wang, Y.-X. and H.~Xu (2013).
\newblock Noisy sparse subspace clustering.
\newblock In {\em International Conference on Machine Learning}, pp.\  89--97.
  PMLR.

\bibitem[\protect\citeauthoryear{Wu, Chen, and Zhou}{Wu
  et~al.}{2016}]{wu2016online}
Wu, J., Y.~Chen, and S.~Zhou (2016).
\newblock Online detection of steady-state operation using a
  multiple-change-point model and exact bayesian inference.
\newblock {\em IIE transactions\/}~{\em 48\/}(7), 599--613.

\bibitem[\protect\citeauthoryear{Wu, Xu, Zhang, and Yuan}{Wu
  et~al.}{2019}]{wu2019sequential}
Wu, J., H.~Xu, C.~Zhang, and Y.~Yuan (2019).
\newblock A sequential bayesian partitioning approach for online steady-state
  detection of multivariate systems.
\newblock {\em IEEE Transactions on Automation Science and Engineering\/}~{\em
  16\/}(4), 1882--1895.

\bibitem[\protect\citeauthoryear{Yuan and Lin}{Yuan and
  Lin}{2006}]{yuan2006model}
Yuan, M. and Y.~Lin (2006).
\newblock Model selection and estimation in regression with grouped variables.
\newblock {\em Journal of the Royal Statistical Society: Series B (Statistical
  Methodology)\/}~{\em 68\/}(1), 49--67.

\bibitem[\protect\citeauthoryear{Zhang, Yan, Lee, and Shi}{Zhang
  et~al.}{2020}]{zhang2020dynamic}
Zhang, C., H.~Yan, S.~Lee, and J.~Shi (2020).
\newblock Dynamic multivariate functional data modeling via sparse subspace
  learning.
\newblock {\em Technometrics\/}, 1--14.

\bibitem[\protect\citeauthoryear{Zhang, Li, Zhang, and Guo}{Zhang
  et~al.}{2016}]{zhang2016low}
Zhang, J., C.-G. Li, H.~Zhang, and J.~Guo (2016).
\newblock Low-rank and structured sparse subspace clustering.
\newblock In {\em 2016 Visual Communications and Image Processing (VCIP)}, pp.\
   1--4. IEEE.

\bibitem[\protect\citeauthoryear{Zhou, Lafferty, and Wasserman}{Zhou
  et~al.}{2010}]{zhou2010time}
Zhou, S., J.~Lafferty, and L.~Wasserman (2010).
\newblock Time varying undirected graphs.
\newblock {\em Machine Learning\/}~{\em 80\/}(2), 295--319.

\bibitem[\protect\citeauthoryear{Zhu, Strawn, and Dunson}{Zhu
  et~al.}{2016}]{zhu2016bayesian}
Zhu, H., N.~Strawn, and D.~B. Dunson (2016).
\newblock Bayesian graphical models for multivariate functional data.

\end{thebibliography}
\end{document}